\documentclass[10pt]{article}
\usepackage[accepted]{tmlr}

\usepackage{amsmath,amsfonts,bm}

\def\eqref#1{equation~\ref{#1}}

\def\1{\bm{1}}

\def\rvr{{\mathbf{r}}}

\def\rvx{{\mathbf{x}}}

\def\rvz{{\mathbf{z}}}

\DeclareMathAlphabet{\mathsfit}{\encodingdefault}{\sfdefault}{m}{sl}
\SetMathAlphabet{\mathsfit}{bold}{\encodingdefault}{\sfdefault}{bx}{n}

\def\gA{{\mathcal{A}}}

\def\gE{{\mathcal{E}}}

\def\gN{{\mathcal{N}}}

\def\gP{{\mathcal{P}}}

\def\gY{{\mathcal{Y}}}

\newcommand{\E}{\mathbb{E}}

\newcommand{\R}{\mathbb{R}}

\newcommand{\KL}{D_{\mathrm{KL}}}

\usepackage{microtype}
\usepackage{graphicx}
\usepackage{booktabs}
\usepackage{hyperref}
\usepackage{float}
\usepackage{url}
\usepackage{xcolor}

\usepackage{amsmath}
\usepackage{amssymb}
\usepackage{mathtools}
\usepackage{amsthm}
\usepackage{multirow}
\usepackage{siunitx}
\usepackage{subcaption}

\theoremstyle{plain}

\theoremstyle{definition}

\theoremstyle{remark}

\usepackage{thmtools}
\usepackage{thm-restate}
\usepackage[capitalize,noabbrev]{cleveref}

\DeclareMathOperator{\enc}{Enc}
\DeclareMathOperator{\proj}{Proj}
\DeclareMathOperator{\dec}{Dec}

\DeclarePairedDelimiter\abs\lvert\rvert
\DeclarePairedDelimiter\norm\lVert\rVert
\DeclarePairedDelimiterX{\KLutil}[2]{(}{)}{#1\;\delimsize\|\;#2}
\renewcommand{\KL}{D_{KL}\KLutil}

\newcommand{\supcon}{\mathcal{L}^{\text{sup}}}
\newcommand{\inv}{\mathcal{L}^{\text{inv}}}

\title{Supervised Contrastive Learning of Environment-Invariant Representations\thanks{Codex was used to assist with editorial revisions; the authors reviewed and approved the final text.}}

\author{
\name Taro Makino \email{taro@nyu.edu} \\
\addr Center for Data Science, New York University
\AND
\name Ji Won Park \email{parj2@gene.com} \\
\addr Prescient Design, Genentech
\AND
\name Natasa Tagasovska \email{natasa.tagasovska@roche.com} \\
\addr Prescient Design, Genentech
\AND
\name Takamasa Kudo \email{kudo.takamasa@gene.com} \\
\addr Research and Early Development (gRED), Genentech
\AND
\name Paula Coelho \email{coelho.paula@gene.com} \\
\addr Research and Early Development (gRED), Genentech
\AND
\name Jan-Christian Huetter \email{huettej1@gene.com} \\
\addr Biology Research | AI Development (BRAID), Genentech
\AND
\name Heming Yao \email{yao.heming@gene.com} \\
\addr Biology Research | AI Development (BRAID), Genentech
\AND
\name Burkhard Hoeckendorf \email{hoeckendorf.burkhard@gene.com} \\
\addr Biology Research | AI Development (BRAID), Genentech
\AND
\name Ana Carolina Leote \email{leote.carolina@gene.com} \\
\addr Research and Early Development (gRED), Genentech
\AND
\name Stephen Ra \email{ras3@gene.com} \\
\addr Prescient Design, Genentech
\AND
\name David Richmond \email{richmond.david@gene.com} \\
\addr Biology Research | AI Development (BRAID), Genentech
\AND
\name Kyunghyun Cho \email kyunghyun.cho@nyu.edu \\
\addr Center for Data Science, New York University \\
\addr Prescient Design, Genentech
\AND
\name Aviv Regev \email{regev.aviv@gene.com} \\
\addr Research and Early Development (gRED), Genentech
\AND
\name Romain Lopez \email{romain.lopez@nyu.edu} \\
\addr Department of Computer Science \& Biology, New York University \\
\addr Research and Early Development (gRED), Genentech
}

\def\month{11}
\def\year{2026}
\def\openreview{\url{https://openreview.net/forum?id=CROGDBt1my}}

\begin{document}

\maketitle

\begin{abstract}
Real-world datasets often combine data collected under different experimental conditions. This yields larger datasets, but can also introduce environment-associated correlations that make it difficult to model the phenomena of interest. We address this with Supervised Contrastive Invariance Learning (SCIL), which learns a task-supervised representation $\rvz_c$ while regularizing its environment-related structure, together with an environment-supervised auxiliary representation $\rvz_s$. The contribution of the auxiliary representation to the reported results remains unresolved. SCIL is based purely on Supervised Contrastive Learning and uses a single hyperparameter $\alpha$ to control the strength of environment regularization. We study this regularization under distribution shift with Colored MNIST and Camelyon17-WILDS, where increasing $\alpha$ decreases in-distribution validation accuracy and increases held-out test accuracy. We also evaluate SCIL on batch correction for 26 million Optical Pooled Screening images, where $\rvz_c$ preserves biological signal while becoming less predictive of inter-well variation.
\end{abstract}

\section{Introduction}
\label{section:introduction}

Real-world machine learning (ML) datasets often combine data collected under different experimental conditions, such as medical images or stained histopathology sections collected at different hospitals~\citep{bandi2019detection,mckinney2020international}. This practice yields larger datasets, but the different experimental conditions alter the images' appearance, and induce spurious correlations that make it difficult to model the phenomena of interest. While human perception is relatively robust~\citep{makino2022differences}, ML models tend to rely on hospital-specific spurious correlations, and fail to generalize out-of-distribution to unseen hospitals~\citep{koh2021wilds}.

Similar spurious correlations are a long-standing problem in experimental biology~\citep{chandrasekaran2024three}, where they are called batch effects~\citep{leek2010tackling}. They can arise between experiments conducted in different labs, within the same lab, and even within a single large parallelized experiment. Removing batch effects by batch correction is an active research direction~\citep{arevalo2024evaluating}.

In some cases, we can manually remove the spurious correlations by using our prior knowledge. For example, color-based data augmentation can remove the staining variation in histopathology images~\citep{nguyen2023contrimix}. Similarly, in experimental biology, there are post-processing methods that remove specific known batch effects~\citep{carpenter2006cellprofiler}. However, such approaches have two significant limitations. First, they require manual post-hoc quality checks to ensure the post-processing did not remove desirable information. Second, some spurious correlations may be unknown, and therefore remain uncorrected. This motivates the development of automated approaches that maximize the removal of the spurious correlations, while minimizing the impact on the phenomena of interest.

To address these issues, we use the observed target $y$ and environment $e$ to train two representations. Let $\rvx \in \R^{D_\rvx}$ be an observation, such as a histopathology image; let $y \in \mathbb{Z}_{\geq 0}$ denote the target, such as disease status; and let $e \in \mathbb{Z}_{\geq 0}$ denote the observed environment, such as the hospital that processed the image. We learn $\rvz_c \in \R^{D_{\rvz_c}}$ with target supervision and environment regularization, and train the auxiliary representation $\rvz_s \in \R^{D_{\rvz_s}}$ with environment supervision. In some experiments, $\rvz_s$ is supervised by the pair $(y,e)$. As we show in Section~\ref{section:theory}, the objective regularizes a first-moment comparison between the same- and different-environment groups in $\rvz_c$, but does not impose a statistical-independence constraint between the representations.

\begin{figure}[htb]
    \centering
    \includegraphics[width=0.5\textwidth]{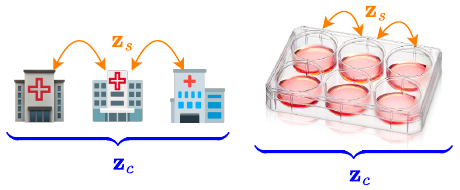}
    \caption{Environment-associated correlations can emerge when collecting medical images from different hospitals or conducting single-cell perturbation screens across multiple wells. SCIL trains $\rvz_c$ with target supervision and environment regularization, and trains the auxiliary representation $\rvz_s$ with environment-associated supervision, using either $e$ or $(y,e)$.}
\end{figure}

Reducing environment-associated structure in a task representation can be useful when that structure reflects a shortcut, but the appropriate amount of regularization is application dependent. Hyperparameter selection under distribution shift is difficult in practice~\citep{gulrajani2020search}, and observed environment information is not necessarily spurious. SCIL therefore requires environment labels during training and requires the user to justify suppressing information associated with the chosen environment variable.

Our primary contribution is Supervised Contrastive Invariance Learning (SCIL), an environment-regularized representation learning method based purely on Supervised Contrastive Learning (SCL)~\citep{khosla2020supervised}. Following the authors' notation, we learn two encoder networks $\enc_c(\cdot)$ and $\enc_s(\cdot)$ that map $\rvx$ to intermediate representations:
\begin{align*}
    \rvr_c &\coloneqq \enc_c(\rvx) \in \R^{D_{\rvr_c}}, \\ 
    \rvr_s &\coloneqq \enc_s(\rvx) \in \R^{D_{\rvr_s}}.
\end{align*}
We additionally learn two projection networks $\proj_c(\cdot)$ and $\proj_s(\cdot)$ that map the intermediate representations to the lower-dimensional embeddings given by
\begin{align*}
    \rvz_c &\coloneqq \proj_c(\rvr_c) \in \R^{D_{\rvz_c}}, \\
    \rvz_s &\coloneqq \proj_s(\rvr_s) \in \R^{D_{\rvz_s}},
\end{align*}
which are normalized to the unit hypersphere. In \citet{khosla2020supervised}, it was shown that some prediction tasks benefit from using the intermediate representations, rather than the embeddings. The representation-learning objective consists of three terms:
\begin{align}
    \label{eq:SCIL}
    \min \supcon_{\rvz_c, y} + \supcon_{\rvz_s, e} + \alpha \inv_{\rvz_c, e}.
\end{align}
The first term clusters $\rvz_c$ by $y$, the second clusters $\rvz_s$ by $e$, and the third is the environment-regularization loss applied to $\rvz_c$. The hyperparameter $\alpha \in \R_{\geq 0}$ controls its strength. Section~\ref{section:theory} characterizes this term: it controls a first-moment quantity comparing each anchor with the mean embeddings of its same- and different-environment groups. Notably, it does not establish statistical independence between $\rvz_c$ and $e$.

We evaluate SCIL on three datasets spanning distribution shift and biological batch correction. On CMNIST and Camelyon17-WILDS~\citep{koh2021wilds}, increasing $\alpha$ decreases in-distribution validation accuracy and increases held-out test accuracy. The value $\alpha=192$ was selected after examining held-out test performance and therefore characterizes this observed trade-off rather than the result of test-independent model selection.

For batch correction, we apply SCIL to more than 26 million images of individual cells~\citep{funk2022phenotypic}. The cells receive \num{5050} genetic perturbations labeled by $y$ and are collected across 34 wells labeled by $e$. Relative to the reported baselines, $\rvz_c$ preserves more signal under the CORUM evaluation while yielding lower well-prediction F1.

\section{Supervised Contrastive Invariance Learning}
\label{section:SCIL}

The three terms in Equation~\ref{eq:SCIL} are defined as follows. Let $I$ index the examples in a minibatch, let $B\coloneqq\abs{I}$ denote the batch size, and let $\gA(i)=I\setminus\{i\}$. Following \citet{khosla2020supervised}, examples are compared by dot product. The environment-regularization loss uses the following log matching probability, whose softmax denominator includes every minibatch position, including the anchor itself:
\begin{align*}
    \log P(M_{i,c}^j=1)=\log\frac{\exp(\rvz_c^i\cdot\rvz_c^j/\tau)}{\sum_{a\in I}\exp(\rvz_c^i\cdot\rvz_c^a/\tau)},\qquad \tau=0.1.
\end{align*}
The projection networks unit-normalize $\rvz_c$ and $\rvz_s$ before the losses are evaluated. The analogous definition for $M_{i,s}^j$ uses $\rvz_s$.

The first term in Equation~\ref{eq:SCIL} is a direct application of SCL, and is given by
\begin{align*}
    \supcon_{\rvz_c, y} = -\frac{1}{B}\sum_{i \in I} \frac{1}{\abs{\gP_y(i)}} \sum_{p \in \gP_y(i)} \log P(M_{i, c}^p = 1),
\end{align*}
where $\gP_y(i) = \{j \in \gA(i) : y^i = y^j\}$ are the positive pairs for the anchor point $i$ with respect to $y$. This is the negative log matching probability averaged over positive pairs and anchors. Minimizing this loss clusters $\rvz_c$ with respect to $y$.

The second term in Equation~\ref{eq:SCIL} is also a direct application of SCL, and is given by
\begin{align*}
    \supcon_{\rvz_s, e} = -\frac{1}{B}\sum_{i \in I} \frac{1}{\abs{\gP_e(i)}} \sum_{p \in \gP_e(i)} \log P(M_{i, s}^p = 1),
\end{align*}
where $\gP_e(i) = \{j \in \gA(i) : e^i = e^j\}$ are the positive pairs for the anchor point $i$ with respect to $e$. Minimizing this loss clusters $\rvz_s$ with respect to $e$. As we later discuss in our experiments (Section~\ref{section:domain generalization}), it can be useful to let $\rvz_s$ represent the variation of $\rvx$ with respect to the pair $(y, e)$, rather than just $e$. We can do this by replacing $\supcon_{\rvz_s, e}$ with
\begin{align*}
    \supcon_{\rvz_s, (y, e)} = -\frac{1}{B}\sum_{i \in I} \frac{1}{\abs{\gP_{(y, e)}(i)}} \sum_{p \in \gP_{(y, e)}(i)} \log P(M_{i, s}^p = 1),
\end{align*}
where $\gP_{(y, e)}(i) = \{j \in \gA(i) : y^i = y^j, e^i = e^j\}$ are the positive pairs for the anchor point $i$ with respect to the pairs of labels $(y, e)$.

The third term is the environment-regularization loss. Define $\gP_e(i)=\{j\in\gA(i):e^i=e^j\}$ and $\gN_e(i)=\{j\in\gA(i):e^i\ne e^j\}$. We define this loss as
\begin{align*}
    \inv_{\rvz_c,e}=\frac{1}{B}\sum_{i\in I}\abs*{\frac{1}{B}\sum_{p\in\gP_e(i)}\log P(M_{i,c}^p=1)-\frac{1}{B}\sum_{q\in\gN_e(i)}\log P(M_{i,c}^q=1)}.
\end{align*}
Appendix~\ref{appendix:code} provides detailed pseudocode for the implementation. Section~\ref{section:theory} analyzes what this objective enforces, including the role of the group cardinalities.

The decoder is not part of Equation~\ref{eq:SCIL}. After representation learning, we hold $\rvz_c$ and $\rvz_s$ fixed and optimize only the decoder with
\begin{align*}
    \min_{\dec}-\log p(\rvx\mid\dec(\rvz_c,\rvz_s)).
\end{align*}
This optional post-hoc objective supports the qualitative CMNIST analysis in Section~\ref{section:qualitative results}; reconstruction does not train the encoders or establish independence between the representations.

\section{Analysis of the environment-regularization loss}
\label{section:theory}

Throughout this section we assume that $\rvz_c$ is unit-normalized, and we abbreviate $\ell_{ij}\coloneqq\log P(M_{i,c}^j=1)$. Note that $\ell_{ij}<0$ for all $i,j$, because the softmax denominator retains the self-similarity term. Fix an anchor $i \in I$ whose two groups are nonempty, write $n^+_i\coloneqq\abs{\gP_e(i)}$ and $n^-_i\coloneqq\abs{\gN_e(i)}$, so that $n^+_i+n^-_i=B-1$, and let $\bar\ell^{\,+}_i$ and $\bar\ell^{\,-}_i$ denote the average of $\ell_{ij}$ over $\gP_e(i)$ and over $\gN_e(i)$ respectively.

We summarize the anchor by three quantities, the \emph{environment gap}, the \emph{scale}, and the \emph{group-size imbalance},
\begin{align*}
    \delta_i\coloneqq\bar\ell^{\,-}_i-\bar\ell^{\,+}_i,\qquad
    \Lambda_i\coloneqq-\tfrac{1}{2}\bigl(\bar\ell^{\,+}_i+\bar\ell^{\,-}_i\bigr)>0,\qquad
    \rho_i\coloneqq\frac{n^-_i-n^+_i}{B-1}\in[-1,1].
\end{align*}
The sign of $\delta_i$ indicates the relative strength of matches within and across environments. For example, a negative $\delta_i$ means that same-environment examples are matched more strongly than different-environment ones. The sign of $\rho_i$ indicates the skewness of the group sizes. For example, with $K$ equally represented environments, $\rho_i$ is positive for $K>2$ and exactly zero for $K=2$. The scale $\Lambda_i$ is the average of the two group means, and is always positive.

The following proposition expresses the loss in terms of the three quantities above. Proofs for this section are given in Appendix~\ref{appendix:proofs}.
\begin{restatable}[Characterization of the environment-regularization loss]{proposition}{propchar}
\label{prop:zero}
If every anchor has nonempty groups, then
\begin{align*}
    \inv_{\rvz_c,e}=\frac{B-1}{2B^2}\sum_{i\in I}\abs*{\delta_i-2\rho_i\Lambda_i}.
\end{align*}
In particular, the loss vanishes if and only if $\delta_i=2\rho_i\Lambda_i$ for every anchor $i$, which for equally sized groups reduces to $\delta_i=0$.
\end{restatable}

Because the two masked sums are divided by $B$, rather than by their respective group sizes, the loss does not generally drive the gap between the group-mean log matching probabilities to zero. If $n_i^+\ne n_i^-$, then $\rho_i\ne0$ and $\Lambda_i>0$, so zero loss requires $\delta_i=2\rho_i\Lambda_i\ne0$. This includes batches with more than two equally represented environments, for which the same- and different-environment groups have unequal sizes.

To see what a positive target asks for, note that $\ell_{ij}=\rvz_c^i\cdot\rvz_c^j/\tau-L_i$, where the log-partition $L_i$ does not depend on $j$. Averaging over each group and subtracting therefore cancels $L_i$, leaving
\begin{align*}
    \delta_i=\frac{1}{\tau}\,\rvz_c^i\cdot\bigl(\mu^-_i-\mu^+_i\bigr),
\end{align*}
where $\mu^+_i$ and $\mu^-_i$ are the mean embeddings of the same- and different-environment groups. The environment gap is thus a first-moment quantity: it compares the anchor against the direction separating its two group centroids. A representation that encodes the environment places same-environment examples closer to the anchor, making $\delta_i$ negative. When $\rho_i>0$, the positive target asks each anchor to match its different-environment neighbors more strongly than its own. The loss therefore constrains an anchor-wise first-moment comparison, but neither equates the group-mean embeddings nor establishes statistical independence from the environment.

\section{Comparison with variational approaches}
\label{section:iVAE}

As a basis of comparison for SCIL, we develop a two-representation algorithm based on Identifiable Variational Autoencoders (iVAEs)~\citep{khemakhem2020variational}. VAEs are popular for modeling single-cell omics data~\citep{wang2023multi,tu2024supervised,mao2024learning}, and several extensions regularize dependence on auxiliary variables~\citep{louizos2016variational,lopez2018information}. We use the iVAE as an empirical comparator; its identifiability framework does not apply to the SCIL objective.

We specify an iVAE with latent representations corresponding to $\rvz_c$ and $\rvz_s$. The generative model is
\begin{align*}
    p_\theta(\rvx, \rvz_c, \rvz_s \mid y, e) = p_\theta(\rvx \mid \rvz_c, \rvz_s) p_\theta(\rvz_c \mid y) p_\theta(\rvz_s \mid e),
\end{align*}
while the inference model is defined as
\begin{align*}
    q_\phi(\rvz_c, \rvz_s \mid \rvx, e) = q_\phi(\rvz_c \mid \rvx) q_\phi(\rvz_s \mid \rvx, e).
\end{align*}
We fit this model by maximizing the evidence lower bound (ELBO)~\citep{jordan1999introduction}, given by
\begin{align*}
    \min_{\theta, \phi} \ &\E_{q_\phi(\rvz_c \mid \rvx) q_\phi(\rvz_s \mid \rvx, e)}[-\log p_\theta(\rvx \mid \rvz_c, \rvz_s)]\\
    + \ &\KL{q_\phi(\rvz_c \mid \rvx)}{p_\theta(\rvz_c \mid y)} \\
    + \ &\KL{q_\phi(\rvz_s \mid \rvx, e)}{p_\theta(\rvz_s \mid e)}.
\end{align*}

Empirically, conditioning the posterior of $\rvz_s$ on both $\rvx$ and $e$, rather than just $\rvx$, significantly impacts downstream performance. We hypothesize that conditioning the posterior of $\rvz_s$ on $e$ makes it easier to encode the variation with respect to $e$ in $\rvz_s$, which reduces the incentive to encode it in $\rvz_c$. We use a mixture of experts approach to condition on $e$, where a neural network takes in $\rvx$ and outputs separate posterior parameters for each value of $e$.

In our evaluations, iVAE performs better than some baselines but worse than SCIL. VAE training must balance reconstruction and KL divergence and can encounter posterior collapse, prior collapse, and numerical instability~\citep{bowman2015generating,razavi2019preventing,fu2019cyclical,dai2020usual,wang2021posterior,child2020very}. We treat these results as an empirical comparison under the reported setup rather than evidence of a general advantage over variational methods.

\section{Experiments}
\label{section:experiments}

We evaluate SCIL on three datasets spanning distribution shift and biological batch correction. Under distribution shift, we use the intermediate representation $\rvr_c$ to predict $y$ in a held-out environment and measure how environment regularization changes in-distribution and out-of-distribution accuracy. Batch correction uses the projection embedding $\rvz_c$ and evaluates CORUM protein-complex prediction and a linear well-label probe. Together, these evaluations characterize target utility and sensitivity to the observed environment.

\subsection{Controlling the in-distribution/out-of-distribution trade-off}
\label{section:domain generalization}

\subsubsection{Evaluation under distribution shift}

We consider data collected in environments representing different conditions, such as the hospital that collected a sample. Let $p_{\text{all}} = \{p_e(\rvx_e,y_e):e\in\gE_{\text{all}}\}$ be a family of distributions indexed by environment. The training data are drawn from environments $\gE_{\text{tr}}$, and evaluation data are drawn from disjoint held-out environments $\gE_{\text{te}}$. We use this setup to measure how the user-specified environment-regularization strength $\alpha$ moves predictive accuracy between in-distribution validation data and a held-out environment. The experiment characterizes this trade-off; it does not provide a test-independent rule for choosing the point on the trade-off for an unknown deployment shift.

\subsubsection{Datasets}

\begin{figure*}[htb]
    \centering
    \hspace*{\fill}
    \begin{subfigure}[t]{0.6\textwidth}
        \centering
        \includegraphics[width=\textwidth]{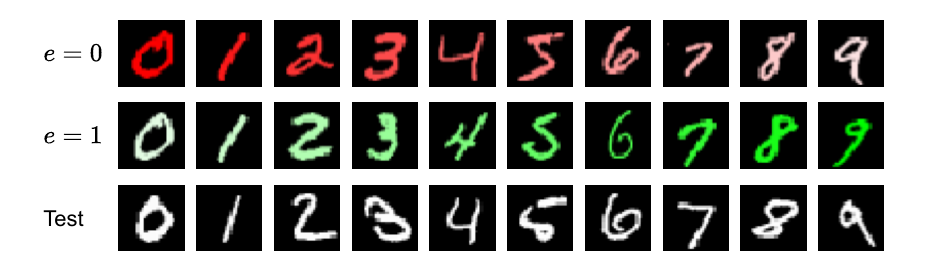}
        \caption{Dataset}
        \label{fig:cmnist}
    \end{subfigure}
    \hfill
    \begin{subfigure}[t]{0.3\textwidth}
        \centering
        \includegraphics[width=0.65\textwidth]{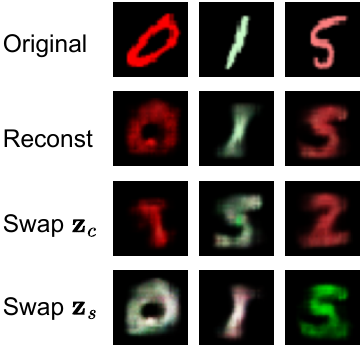}
        \caption{Counterfactual generation}
        \label{fig:cmnist,gen,scbd}
    \end{subfigure}
    \hspace*{\fill}
    \vspace{-0.2cm}
    \caption{Colored MNIST. (a) The training data contain an environment-dependent correlation between color and digit; every test digit is white. (b) A post-hoc decoder qualitatively visualizes swaps of $\rvz_c$ and $\rvz_s$. Swapping $\rvz_c$ changes the decoded digit in these examples, while swapping $\rvz_s$ changes the decoded color.}
    \vspace{-0.2cm}
\end{figure*}

In addition to Camelyon17-WILDS, we use Colored MNIST (CMNIST), extending \citet{arjovsky2019invariant}. The target $y\in\{0,\dotsc,9\}$ is the digit. Two training environments have environment-dependent correlations between color and digit, while every test digit is white (Figure~\ref{fig:cmnist}). ERM exhibits a negative correlation between in-distribution and out-of-distribution performance (Appendix Figure~\ref{fig:cmnist,val_test_scatter}). We train $\rvz_c$ with digit supervision and environment regularization and train $\rvz_s$ with joint $(y,e)$ supervision.

Camelyon17-WILDS~\citep{koh2021wilds} is a patch-based histopathology dataset for binary tumor classification, collected across five hospitals~\citep{bandi2019detection}, three of which form the training environments. Hospital-associated staining differences can create shortcuts~\citep{tellez2019quantifying}. We train $\rvz_c$ with tumor-label supervision and hospital regularization and train $\rvz_s$ with joint $(y,e)$ supervision. The reported study does not directly measure held-out hospital prediction from $\rvz_c$.

\subsubsection{Baselines}

We report ERM, CORrelation ALignment (CORAL)~\citep{sun2016deep}, Domain-Adversarial Neural Networks (DANN)~\citep{ganin2016domain}, Invariant Risk Minimization (IRM)~\citep{arjovsky2019invariant}, Fish~\citep{shi2021gradient}, Group Distributionally Robust Optimization (Group DRO)~\citep{sagawa2020distributionally}, and the iVAE from Section~\ref{section:iVAE}. The provenance and protocol differences of these results are described below.

\subsubsection{Qualitative results}
\label{section:qualitative results}

Figure~\ref{fig:cmnist,gen,scbd} is a qualitative diagnostic produced by the post-hoc decoder. In the displayed examples, swapping $\rvz_c$ changes the decoded digit while swapping $\rvz_s$ changes the decoded color. The decoder is trained after representation learning with the representations fixed. Appendix Figure~\ref{fig:cmnist,umap} directly compares the CMNIST validation embeddings at $\alpha=0$ and $\alpha=192$, colored by digit and environment. The $\rvz_c$-only ablations in Appendix Figures~\ref{fig:cmnist,resnet18,single_block} and~\ref{fig:camelyon17,single_block} show that $\rvz_s$ is not required for the downstream trade-off in every setting.

\subsubsection{Quantitative results}

\begin{figure*}[htb]
    \centering
    \hspace*{\fill}
    \begin{subfigure}[t]{0.4\textwidth}
        \centering
        \includegraphics[width=\textwidth]{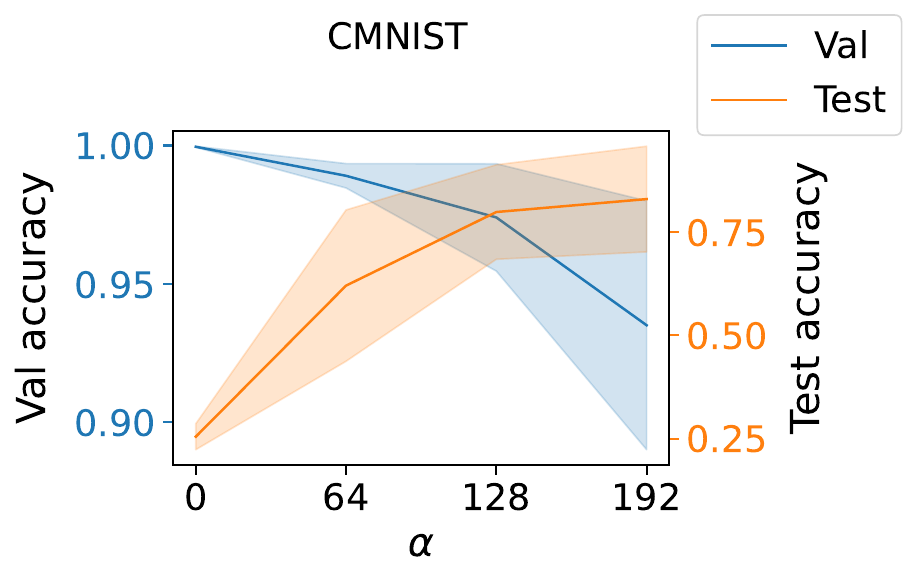}
    \end{subfigure}
    \hfill
    \begin{subfigure}[t]{0.4\textwidth}
        \centering
        \includegraphics[width=\textwidth]{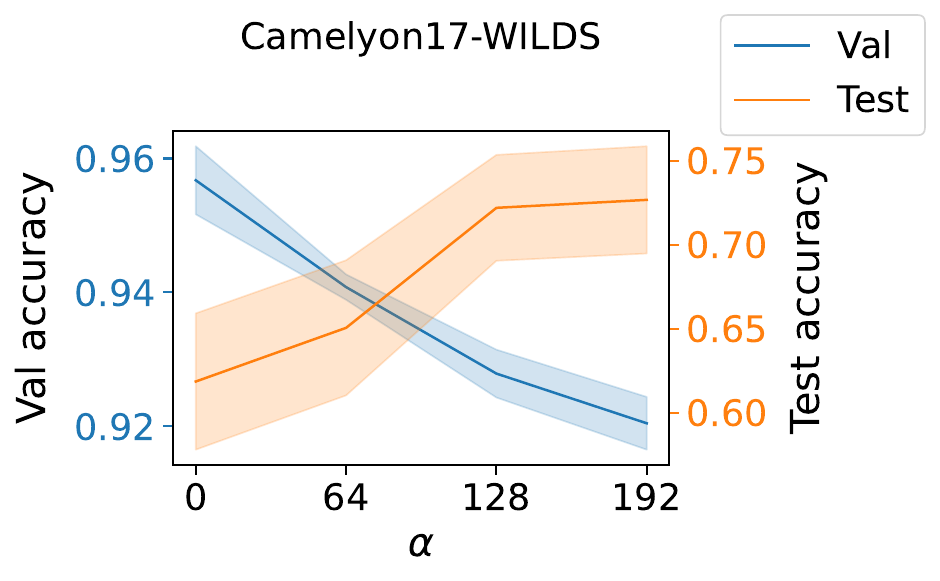}
    \end{subfigure}
    \hspace*{\fill}
    \vspace{-0.2cm}
    \caption{Increasing the environment-regularization strength $\alpha$ decreases validation accuracy and increases test accuracy on the two reported shifts.}
    \vspace{-0.2cm}
    \label{fig:domain generalization}
\end{figure*}

Figure~\ref{fig:domain generalization} shows the effect of increasing the environment-regularization strength $\alpha$: validation accuracy decreases while held-out test accuracy increases on CMNIST and Camelyon17-WILDS.

\begin{table}[htb]
    \centering
    \caption{Contextual test accuracy (\%) over ten random seeds. The SCIL rows show endpoints of the environment-regularization trade-off. The value $\alpha=192$ was selected after examining held-out test performance; comparator protocols are not uniformly matched.}
    \begin{tabular}{lcc}
    \toprule
    Algorithm & CMNIST & Camelyon17-WILDS\\
    \midrule
    SCIL ($\alpha = 0$) & $25.5 \pm 3.0$ & $61.9 \pm 3.8$\\
    SCIL ($\alpha = 192$) & $82.9 \pm 12.1$ & $72.7 \pm 3.0$\\ \midrule
    ERM & $37.8 \pm 2.6$ & $65.8 \pm 4.9$ \\ 
    CORAL & $37.6 \pm 3.6$ & $59.5 \pm 7.7$\\
    DANN & $39.0 \pm 4.5$ & $55.2 \pm 6.7$ \\
    IRM & $37.0 \pm 4.2$ & $66.3 \pm 2.1$\\
    Fish & $48.2 \pm 3.5$ & $49.1 \pm 0.9$\\
    Group DRO & $35.0 \pm 2.9$ & $68.4 \pm 7.3$\\ \midrule
    iVAE & $52.1 \pm 37.6$ & $52.0 \pm 2.0$\\
    \bottomrule
    \end{tabular}
    \label{table:domain_generalization}
\end{table}

Table~\ref{table:domain_generalization} provides contextual test accuracies. It reports ordinary classification accuracy, rather than worst-group or macro accuracy, with mean and standard deviation over ten seeds. For SCIL, Camelyon17-WILDS images are resized to $32\times32$, the main result uses DenseNet-121 encoders, training uses batches of 2,048 for at most 25,000 steps, and the checkpoint minimizing validation loss is selected. The appendix also reports ResNet-18 results. The encoders use randomly initialized weights, and we apply no data augmentation. For the baseline algorithms, hyperparameters are selected using in-distribution validation performance (Appendix Table~\ref{table:hyperparameter search space}). Most Camelyon17-WILDS baseline values are copied from the WILDS leaderboard; Fish and SCIL are evaluated by us. Common preprocessing, architecture, pretraining, training budget, and validation protocols have not been established for the leaderboard values, precluding direct comparison. For Fish, we include pretraining duration in the hyperparameter search, whereas the published leaderboard value used the duration yielding the best test accuracy~\citep{shi2021gradient}.

For SCIL, applying the same in-distribution selection principle to learning rate, weight decay, and environment-regularization strength selects $\alpha=0$. We also report $\alpha=192$, selected after examining held-out test performance, to characterize the observed decrease in validation accuracy and increase in test accuracy as $\alpha$ increases. This value is not the output of a test-independent model-selection procedure. Practical deployment therefore requires a pre-specified invariance target, domain knowledge, or a test-independent design such as leave-one-training-environment-out validation; none of these selection procedures is validated here.

\subsubsection{Scope and interpretation}

SCIL is intended for settings in which observed environment-associated features create a negative trade-off between in-distribution and out-of-distribution performance and domain knowledge supports suppressing those features. It requires observed environment labels during training. Environment-associated information is not automatically spurious, and users must justify reducing it for the intended downstream use.

We assess the direction of the trade-off by training Empirical Risk Minimization (ERM)~\citep{vapnik1995nature} across a hyperparameter search space and comparing in-distribution and out-of-distribution performance. \citet{teney2024id} found a pronounced negative correlation on Camelyon17-WILDS~\citep{koh2021wilds}. Such a negative relationship is the exception rather than the rule: in a study of 172 datasets, including datasets from DomainBed~\citep{gulrajani2020search} and WILDS~\citep{koh2021wilds}, \citet{wenzel2022assaying} found that in-distribution and out-of-distribution performance are often positively correlated.

PACS~\citep{li2017deeper} and VLCS~\citep{fang2013unbiased} exhibit this positive correlation (Appendix Sections~\ref{appendix:PACS} and~\ref{appendix:VLCS}). Increasing environment regularization does not improve OOD accuracy on these datasets. These negative results delimit the method's scope: SCIL is not expected to help without a shortcut-induced negative ID--OOD trade-off.

\subsection{Batch correction with a real-world Optical Pooled Screen dataset}
\label{section:batch correction}

\subsubsection{Problem description and dataset}

We next evaluate SCIL on batch correction using the Optical Pooled Screen (OPS) dataset~\citep{feldman2019optical,funk2022phenotypic}, which contains 26 million images of single cells receiving one of \num{5050} genetic perturbations, including a non-targeting control. The $100\times100$ images contain four channels measuring DNA damage, F-actin, DNA content, and microtubules. The evaluation asks whether learned representations preserve perturbation-related morphology while becoming less predictive of the well in which a cell was observed.

The dataset was collected in one laboratory across 34 wells, and minor experimental differences can produce well-associated image variation. We train $\rvz_c$ with perturbation supervision and well regularization, and train $\rvz_s$ with well supervision. This statistical training setup does not establish a causal separation of perturbation and well effects.

For each cell image $\rvx$, $y$ denotes the perturbation and $e$ the well. We use ResNet-18 encoders and $D_{\rvz_c}=D_{\rvz_s}=64$. Batch-correction evaluations use the projection embedding $\rvz_c$, while the distribution-shift classifiers use the intermediate representation $\rvr_c$. We report $\alpha=1$ for the OPS evaluation and interpret its effect through CORUM performance and well-label prediction rather than as proof that $\rvz_c$ is independent of $e$. Training uses $64\times64$ images, batches of 2,048 containing 256 perturbation labels and eight examples per label, at most 150,000 steps, three random seeds, and one NVIDIA A100 GPU per run. The separate distribution-shift batch-size ablations show better performance with larger batches; we report no OPS batch-size ablation.

We evaluate two tasks to understand the degree to which we remove the influence of $e$, while preserving the information in $y$. We describe the tasks at a high level here, and provide details in Appendix~\ref{appendix:batch correction}.

The first task is CORUM prediction, which is one measure of the biological information content in the embeddings. This task relies on the CORUM database~\citep{ruepp2010corum} as the ground truth of whether two genes are functionally related based on their membership in the same protein complex (a definition previously used in the context of this biological screen). We take the biological embeddings corresponding to those genetic perturbations $y$, interpret their dot product as the prediction that they are similar, and use these predictions to compute the area under the precision-recall curve. We want the performance on this task to be strong.

The second task is to use the perturbation embeddings to predict $e$, which measures the sensitivity to inter-well batch effects. We fit a linear classifier on top of each of the embeddings, and compute the F1 score. In contrast to the first task, we want the performance on this task to be weak.

\begin{figure*}[htb]
    \centering
    \includegraphics[width=0.7\textwidth]{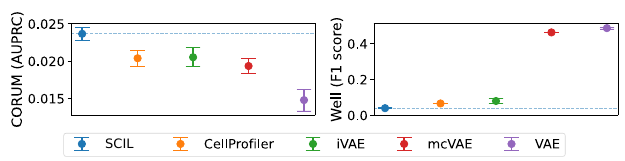}
    \vspace{-0.2cm}
    \caption{Comparison of SCIL with the reported CellProfiler and VAE-based baselines. Left: CORUM protein-complex prediction; higher is better. Right: linear well-label prediction from $\rvz_c$; lower is better.}
    \vspace{-0.2cm}
    \label{fig:funk22,resnet18}
\end{figure*}

\subsubsection{Baselines}

CellProfiler~\citep{carpenter2006cellprofiler} produces manually engineered morphological features that are post-processed for plate and well correction. Following conventional practice, we use the top 64 principal components as the reported CellProfiler representation.

The remaining reported baselines are VAE-based models: the iVAE from Section~\ref{section:iVAE}, Multi-Contrastive VAE (mcVAE)~\citep{wang2023multi}, and a vanilla VAE~\citep{kingma2014auto}. Each VAE latent representation has 64 dimensions. The perturbation representation is $\rvz_c$ for SCIL and iVAE, the salient representation for mcVAE, and the single available representation for CellProfiler and vanilla VAE.

\subsubsection{Quantitative results}

Figure~\ref{fig:funk22,resnet18} shows that SCIL has higher CORUM performance and lower well-prediction F1 than the reported baselines. Thus, $\rvz_c$ preserves more signal under the CORUM metric while being less linearly predictive of well identity. Residual well information beyond the linear probe remains an important consideration for downstream use.

\section{Related work}

\subsection{Relation to disentangled representation learning}

Disentangled representation learning studies representations with independent and semantically meaningful components or groups of components~\citep{higgins2017beta,esmaeili2019structured,kim2018disentangling,chen2018isolating,von2021self,lachapelle2022partial,kong2022partial,lachapelle2024additive,lopez2024toward}. SCIL does not impose an independence penalty or provide an identifiability result. Section~\ref{section:theory} shows that its regularizer controls a first-moment comparison between the same- and different-environment groups in $\rvz_c$, rather than a measure of dependence. We therefore position it as environment-regularized representation learning rather than as a disentanglement method.

\subsection{Invariant representation learning and domain generalization}

The challenge of domain generalization has gained significant attention as ML systems often fail to generalize out-of-distribution. \citet{peters2016causal} introduced a framework for causal inference using invariant prediction, helping maintain predictive accuracy under interventions or environmental changes. Building on this foundation, \citet{arjovsky2019invariant} proposed IRM, a learning paradigm for learning an embedding of the data representation such that the optimal classifier on top of that representation remains invariant across different environments. These works, as well as many extensions~\citep{lu2021invariant}, have been benchmarked on datasets created by the research community, such as those in the DomainBed~\citep{gulrajani2020search} and WILDS~\citep{koh2021wilds} suites. \citet{gulrajani2020search} revealed that with rigorous model selection, ERM often achieves state-of-the-art performance, challenging the perceived benefits of more complex domain generalization methods.

Related methods enforce agreement across source environments at different levels. V-REx penalizes variation in prediction risk across environments~\citep{krueger2021out}, whereas Fishr matches environment-level variances of per-example loss gradients~\citep{rame2022fishr}. SWAD does not impose an invariance criterion; it densely averages model weights over an interval selected using source validation loss to seek a flatter solution~\citep{cha2021swad}. These methods respectively regularize prediction risks, optimization statistics, and the training trajectory. In contrast, SCIL regularizes pairwise matching statistics in the task representation. These approaches all use different criteria and do not by themselves establish that the resulting representation is statistically independent of the environment.

Several domain-generalization methods instead act directly on representation geometry. SelfReg aligns same-class features and logits using positive pairs and same-class feature perturbations~\citep{kim2021selfreg}, while memory-based supervised contrastive learning uses a queue to enlarge the set of same- and different-class examples~\citep{tan2022domain}. PCL argues that directly aligning difficult same-class samples across domains can impair generalization and replaces positive sample-to-sample relations with learned class proxies~\citep{yao2022pcl}. MatchDG further shows that class-conditional domain invariance is not sufficient in general and seeks matches corresponding to the same underlying object across environments~\citep{mahajan2021domain}. More recently, CCFP learns feature perturbations that simulate domain shifts and enforces semantic consistency under those perturbations without using domain labels~\citep{li2023cross}. SCIL differs by explicitly comparing same- and different-environment matching statistics in $\rvz_c$ while training the auxiliary representation $\rvz_s$ with environment labels.

\section{Limitations}

First, our evidence concerns predictive utility and environment sensitivity rather than a complete decomposition of target- and environment-related information. We do not directly measure dependence between $\rvz_c$ and $\rvz_s$, and the $\rvz_c$-only ablations show that the auxiliary representation is not required for the downstream trade-off in every setting. Second, the distribution-shift experiments do not provide a test-independent rule for selecting $\alpha$. The Camelyon17-WILDS comparison also combines results obtained under different protocols, so it should be interpreted as contextual rather than as a state-of-the-art comparison.

\section{Conclusion}

We presented Supervised Contrastive Invariance Learning (SCIL), an SCL-based method that learns a task-supervised representation while regularizing its environment-related structure. On CMNIST and Camelyon17-WILDS, increasing the user-specified regularization strength decreases in-distribution validation accuracy and increases held-out test accuracy. In the OPS experiment, the learned representation achieves higher CORUM performance with lower linear well-prediction F1 than the reported baselines.

SCIL requires the environment label $e$ during training. The appropriate value of $\alpha$ and the decision to suppress environment-associated information require domain knowledge or a test-independent validation design. The post-hoc CMNIST decoder is a toy qualitative analysis and not a validated clinical or biological image-generation system. Future work should directly measure residual target and environment information in both representations, evaluate dependence and minibatch sensitivity, and validate model selection without access to the deployment environment.

\section*{Acknowledgements and Disclosure of Funding}

We thank Henri Dwyer, Joseph Kleinhenz, Philipp Hanslovsky, and Shenglong Wang for data- and compute-related help with this project, and the Genentech and NYU high-performance computing teams for helping us access the compute resources that were used in this work.

This work was done while Taro Makino was an intern at Genentech. Natasa Tagasovska, Takamasa Kudo, Paula Coelho, Heming Yao, Burkhard Hoeckendorf, Ana Carolina Leote, David Richmond, Kyunghyun Cho, and Aviv Regev are employees of Genentech. Ji Won Park, Jan-Christian Huetter, Stephen Ra, and Romain Lopez were affiliated with Genentech during this work. Ji Won Park, Natasa Tagasovska, Jan-Christian Huetter, Heming Yao, Burkhard Hoeckendorf, Stephen Ra, David Richmond, Kyunghyun Cho, and Aviv Regev have equity in Roche. Aviv Regev is a co-founder and equity holder of Celsius Therapeutics and an equity holder in Immunitas. She was an SAB member of ThermoFisher Scientific, Syros Pharmaceuticals, Neogene Therapeutics, and Asimov until July 31, 2020, and has been an employee of Genentech since August 1, 2020. Kyunghyun Cho is supported by the Samsung Advanced Institute of Technology (under the project Next Generation Deep Learning: From Pattern Recognition to AI). Romain Lopez is supported by a Schmidt Sciences AI2050 Fellowship Award. 

\section*{Broader Impact Statement}

Environment-associated shortcuts can undermine medical and biological models. For example, image models can rely on acquisition artifacts or treatment-related features rather than the pathology of interest~\citep{makino2022differences,winkler2019association,oakden2020hidden}. At the same time, an environment label such as hospital identity can correlate with clinically meaningful differences in patient populations, acquisition protocols, disease prevalence, or demographic groups. Reducing this information can remove useful signal or harm particular subgroups. Similarly, an incorrect separation of well-associated and perturbation-associated variation can yield misleading conclusions about gene function or morphology. Environment-associated variation must therefore be assessed rather than automatically treated as nuisance. Applications require validation across clinically and biologically relevant subgroups, domain-expert review, and independent confirmation. SCIL representations should not be the sole basis for clinical, biological, or therapeutic decisions.

The post-hoc decoder in this paper is a toy CMNIST visualization, not a validated medical image generator. Extending counterfactual generation to realistic biomedical images would create risks of fabrication, misleading interpretation, and dual use. Such systems would require provenance controls, explicit labeling of synthetic outputs, expert oversight, and task-specific safety validation.

\bibliography{main}
\bibliographystyle{tmlr}

\newpage
\appendix
\onecolumn

\section*{Appendix}

\section{Proofs}
\label{appendix:proofs}

We restate the notation of Section~\ref{section:theory}. Let $I$ index a minibatch of size $B=\abs{I}$, and for $i\in I$ let $\gA(i)=I\setminus\{i\}$. The projection network unit-normalizes $\rvz_c$, so $\norm{\rvz_c^j}=1$ for every $j$. Write
\begin{align*}
    \ell_{ij}\coloneqq\log P(M_{i,c}^j=1)=\frac{\rvz_c^i\cdot\rvz_c^j}{\tau}-L_i,
    \qquad L_i\coloneqq\log\sum_{a\in I}\exp\!\left(\frac{\rvz_c^i\cdot\rvz_c^a}{\tau}\right),
\end{align*}
noting that the log-partition $L_i$ depends on the anchor $i$ but not on $j$. The two groups of anchor $i$ are $\gP_e(i)=\{j\in\gA(i):e^i=e^j\}$ and $\gN_e(i)=\{j\in\gA(i):e^i\neq e^j\}$, with sizes $n^+_i=\abs{\gP_e(i)}$ and $n^-_i=\abs{\gN_e(i)}$. Because $\gP_e(i)$ and $\gN_e(i)$ partition $\gA(i)$, we have
\begin{align*}
    n^+_i+n^-_i=\abs{\gA(i)}=B-1 .
\end{align*}
The group averages of the log matching probabilities and the corresponding group-mean embeddings are
\begin{align*}
    \bar\ell^{\,+}_i=\frac{1}{n^+_i}\sum_{p\in\gP_e(i)}\ell_{ip},\quad
    \bar\ell^{\,-}_i=\frac{1}{n^-_i}\sum_{q\in\gN_e(i)}\ell_{iq},\quad
    \mu^+_i=\frac{1}{n^+_i}\sum_{p\in\gP_e(i)}\rvz_c^p,\quad
    \mu^-_i=\frac{1}{n^-_i}\sum_{q\in\gN_e(i)}\rvz_c^q,
\end{align*}
and the environment gap, scale, and group-size imbalance are $\delta_i=\bar\ell^{\,-}_i-\bar\ell^{\,+}_i$, $\Lambda_i=-\tfrac12(\bar\ell^{\,+}_i+\bar\ell^{\,-}_i)$, and $\rho_i=(n^-_i-n^+_i)/(B-1)$. Throughout we assume both groups are nonempty, so that these averages are defined.

\propchar*

\begin{proof}
Recall from Section~\ref{section:SCIL} that the loss is
\begin{align*}
    \inv_{\rvz_c,e}=\frac{1}{B}\sum_{i\in I}\abs*{\frac{1}{B}\sum_{p\in\gP_e(i)}\ell_{ip}-\frac{1}{B}\sum_{q\in\gN_e(i)}\ell_{iq}} .
\end{align*}
\emph{Step 1: rewrite the two masked sums using the group averages.} By definition of $\bar\ell^{\,+}_i$ and $\bar\ell^{\,-}_i$ we have $\sum_{p\in\gP_e(i)}\ell_{ip}=n^+_i\bar\ell^{\,+}_i$ and $\sum_{q\in\gN_e(i)}\ell_{iq}=n^-_i\bar\ell^{\,-}_i$. Pulling the common factor $1/B$ out of the absolute value gives
\begin{align*}
    \inv_{\rvz_c,e}=\frac{1}{B^2}\sum_{i\in I}\abs*{n^+_i\bar\ell^{\,+}_i-n^-_i\bar\ell^{\,-}_i} . \tag{A.1}\label{eq:proof-step1}
\end{align*}
\emph{Step 2: change coordinates from $(\bar\ell^{\,+}_i,\bar\ell^{\,-}_i)$ to $(\delta_i,\Lambda_i)$.} The map defining $\delta_i$ and $\Lambda_i$ is linear and invertible, with inverse
\begin{align*}
    \bar\ell^{\,+}_i=-\Lambda_i-\frac{\delta_i}{2},\qquad
    \bar\ell^{\,-}_i=-\Lambda_i+\frac{\delta_i}{2} ,
\end{align*}
as may be checked by substitution: $\bar\ell^{\,-}_i-\bar\ell^{\,+}_i=\delta_i$ and $-\tfrac12(\bar\ell^{\,+}_i+\bar\ell^{\,-}_i)=\Lambda_i$.

\emph{Step 3: substitute and collect terms.} Using the inverse map,
\begin{align*}
    n^+_i\bar\ell^{\,+}_i-n^-_i\bar\ell^{\,-}_i
    &= n^+_i\left(-\Lambda_i-\frac{\delta_i}{2}\right)-n^-_i\left(-\Lambda_i+\frac{\delta_i}{2}\right)\\
    &= \bigl(n^-_i-n^+_i\bigr)\Lambda_i-\frac{n^+_i+n^-_i}{2}\,\delta_i .
\end{align*}
\emph{Step 4: eliminate the group sizes.} By the partition identity $n^+_i+n^-_i=B-1$, and by the definition of $\rho_i$ we have $n^-_i-n^+_i=\rho_i(B-1)$. Hence
\begin{align*}
    n^+_i\bar\ell^{\,+}_i-n^-_i\bar\ell^{\,-}_i
    =\rho_i(B-1)\Lambda_i-\frac{B-1}{2}\,\delta_i
    =\frac{B-1}{2}\bigl(2\rho_i\Lambda_i-\delta_i\bigr) .
\end{align*}
\emph{Step 5: assemble.} Substituting into~\eqref{eq:proof-step1}, using $\abs{-x}=\abs{x}$ and $B-1>0$,
\begin{align*}
    \inv_{\rvz_c,e}
    =\frac{1}{B^2}\sum_{i\in I}\frac{B-1}{2}\abs*{2\rho_i\Lambda_i-\delta_i}
    =\frac{B-1}{2B^2}\sum_{i\in I}\abs*{\delta_i-2\rho_i\Lambda_i} ,
\end{align*}
which is the stated identity.

\emph{Step 6: the vanishing condition.} For $B\geq2$ the prefactor $(B-1)/(2B^2)$ is strictly positive, and the sum is a finite sum of nonnegative terms. Such a sum is zero if and only if every term is zero, that is, if and only if $\delta_i=2\rho_i\Lambda_i$ for every $i\in I$. If the two groups have equal size then $n^-_i-n^+_i=0$, so $\rho_i=0$ and the condition reduces to $\delta_i=0$.
\end{proof}

\section{Experiments}

\subsection{Experimental setup}
\label{appendix:experimental setup}

Every experiment used a single NVIDIA A100 GPU on our institutions' high-performance computing clusters.

\subsubsection{Supervised Contrastive Invariance Learning}

For SCIL, the temperature is $\tau=0.1$. We optimize with AdamW~\citep{loshchilov2019decoupled}, a learning rate of \num{1e-4}, weight decay of 0.01, and batch size \num{2048}.

For domain generalization, we resize the images to $32 \times 32$ pixels. We set $D_{\rvz_c} = D_{\rvz_s} = 128$, which we adopt from \citet{khosla2020supervised}. We train for a maximum of \num{25000} steps, and select the weights that minimize the validation loss. We obtain error bars by repeating each experiment with ten random seeds.

For batch correction, we resize images to $64\times64$ and set $D_{\rvz_c}=D_{\rvz_s}=64$. Each minibatch contains 256 sampled perturbation labels and eight examples per label, for a total batch size of 2,048. We train for at most \num{150000} steps, select the checkpoint minimizing validation loss, and report three random seeds.

We use standard architectures such as ResNet-18~\citep{he2016deep} and DenseNet-121~\citep{huang2017densely} for the encoders $\enc_c(\rvx)$ and $\enc_s(\rvx)$. The projection networks $\proj_c(\rvr_c)$ and $\proj_s(\rvr_s)$ are two-layer Multilayered Perceptrons~\citep{rumelhart1986learning} with hidden sizes of $D_{\rvr_c}$ and $D_{\rvr_s}$, and GELU activations~\citep{hendrycks2016gaussian}. The post-hoc decoder architecture is shown in Appendix Table~\ref{table:SCIL decoder architecture}. We use an additive decoder~\citep{lachapelle2024additive} and define
\begin{align*}
    \log p(\rvx \mid \rvz_c, \rvz_s) = \dec_c(\rvz_c) + \dec_s(\rvz_s),
\end{align*}
where both $\dec_c$ and $\dec_s$ have the same architecture.

\begin{table}[H]
    \centering
    \caption{SCIL post-hoc decoder architecture}
    \begin{tabular}{l}
    \toprule
    \texttt{Linear(64, 256 * (2 ** 2))}\\
    \texttt{ConvTranspose2d(256, 256, 3, stride=2, padding=1, output\_padding=1)}\\
    \texttt{Conv2d(256, 256, 3, padding=1)}\\
    \texttt{ConvTranspose2d(256, 256, 3, stride=2, padding=1, output\_padding=1)}\\
    \texttt{Conv2d(256, 256, 3, padding=1)}\\
    \texttt{ConvTranspose2d(256, 256, 3, stride=2, padding=1, output\_padding=1)}\\
    \texttt{Conv2d(256, 256, 3, padding=1)}\\
    \texttt{ConvTranspose2d(256, 256, 3, stride=2, padding=1, output\_padding=1)}\\
    \texttt{Conv2d(256, 128, 3, padding=1)}\\
    \texttt{Conv2d(128, img\_ch, 1)}\\
    \bottomrule
    \end{tabular}
    \label{table:SCIL decoder architecture}
\end{table}

\subsubsection{Variational Autoencoders}
\label{appendix:VAE}

For our experiments with VAE-based approaches, we use the same experimental setup used in \citet{wang2023multi}, including the architecture and hyperparameters. We resize the images to $64 \times 64$ pixels and use a batch size of 1024. The encoder and decoder architectures are in Appendix Tables~\ref{table:VAE encoder architecture} and \ref{table:VAE decoder architecture}, with GELU activations~\citep{hendrycks2016gaussian} between layers. Since the CMNIST images are $32 \times 32$, we modify the architectures to reduce the up- and down-sampling. For optimization, we use the AdamW~\citep{loshchilov2019decoupled} optimizer with \num{1e-4} learning rate and 0.01 weight decay. We additionally skip gradients with a norm above \num{1e12}, and clip gradients with a norm above \num{1e6}, as done in \citet{child2020very}. We train for a maximum of \num{50000} steps for domain generalization, and three epochs for batch correction, and select the weights with minimum validation loss. We report the validation and test performance across ten random seeds for domain generalization, and three random seeds for batch correction.

\begin{table}[H]
    \centering
    \caption{VAE encoder architecture}
    \begin{tabular}{l}
    \toprule
    \texttt{Conv2d(img\_c, 32, 3, stride=2, padding=1)}\\
    \texttt{Conv2d(32, 32, 3, padding=1)}\\
    \texttt{Conv2d(32, 64, 3, stride=2, padding=1)}\\
    \texttt{Conv2d(64, 64, 3, padding=1)}\\
    \texttt{Conv2d(64, 64, 3, stride=2, padding=1)}\\
    \texttt{Linear(64 * (8 ** 2), 2 * 64)}\\
    \bottomrule
    \end{tabular}
    \label{table:VAE encoder architecture}
\end{table}

\begin{table}[H]
    \centering
    \caption{VAE decoder architecture}
    \begin{tabular}{l}
    \toprule
    \texttt{Linear(2 * 64, 64 * (8 ** 2))}\\
    \texttt{ConvTranspose2d(64, 64, 3, stride=2, padding=1, out\_padding=1)}\\
    \texttt{Conv2d(64, 64, 3, padding=1)}\\
    \texttt{ConvTranspose2d(64, 32, 3, stride=2, padding=1, out\_padding=1)}\\
    \texttt{Conv2d(32, 32, 3, padding=1)}\\
    \texttt{ConvTranspose2d(32, img\_c, 3, stride=2, padding=1, out\_padding=1)}\\
    \bottomrule
    \end{tabular}
    \label{table:VAE decoder architecture}
\end{table}

\subsubsection{Other baselines}

\begin{table}[H]
    \centering
    \caption{Hyperparameter search space}
    \begin{tabular}{lll}
    \toprule
    Condition & Hyperparameter & Search space\\
    \midrule
    \multirow{4}{*}{CMNIST} & Learning rate & \{0.0001, 0.001, 0.01\}\\
    & Weight decay & \{0, 0.001, 0.01\}\\
    & Batch size & \{32\}\\
    & Maximum epochs & \{1, 20, 100\}\\
    \midrule
    \multirow{4}{*}{Camelyon17-WILDS} & Learning rate & \{0.0001, 0.001, 0.01\}\\
    & Weight decay & \{0, 0.001, 0.01\}\\
    & Batch size & \{32\}\\
    & Maximum epochs & \{5\}\\
    \midrule
    CORAL & Penalty weight & \{0.1, 1, 10\}\\
    \midrule
    DANN & Penalty weight & \{0.1, 1, 10\}\\
    \midrule
    IRM & Penalty weight & \{1, 10, 100, 1000\}\\
    \midrule
    \multirow{2}{*}{Fish} & Pretrain steps & \{1000, 10000\}\\
    & Meta learning rate & \{0.001, 0.01, 0.1\}\\
    \midrule
    Group DRO & Step size & \{0.01\}\\
    \bottomrule
    \end{tabular}
    \label{table:hyperparameter search space}
\end{table}

\subsection{Domain generalization}

\subsubsection{Colored MNIST}
\label{appendix:CMNIST}

\paragraph{Data generating process}

The images are $32 \times 32$ pixels, and are RGB. There are two training environments and a test environment. In the first training environment, which we label $e = 0$, we set the foreground pixels in the red channel to the value one, and those in the green and blue channels to the value $y / \abs{\gY}$, where $\abs{\gY} = 10$ is the number of digits. For images with the digit zero, $y = 0$, so the digit is colored completely red. Then, as the digit increases from zero to nine, the digits are colored red, but with a decreasing intensity. In the second training environment, which we label $e = 1$, we set the foreground pixels in the green channel to the value one, and those in the red and blue channels to the value $(\abs{\gY} - 1 - y) / \abs{\gY}$. This has the effect of the digits being colored green, where the intensity increases with as the digit increases from zero to nine. In the test environment, the foreground pixels are set to one in all channels, which makes all of the digits white.

\begin{figure}[H]
    \centering
    \includegraphics[width=0.82\textwidth]{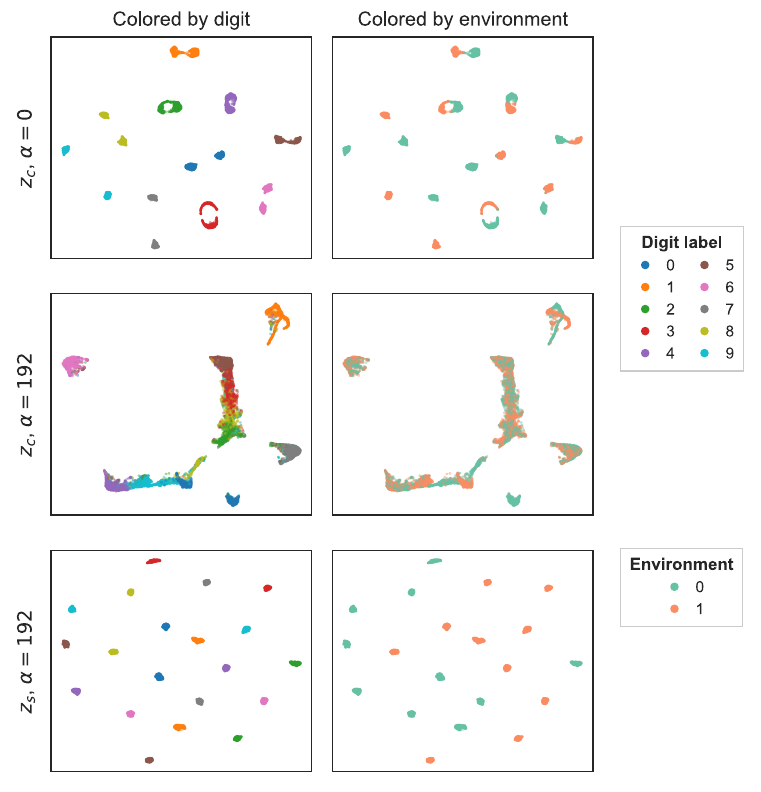}
    \caption{Cosine-UMAP visualization of CMNIST validation embeddings from the $\alpha=0$ and $\alpha=192$, seed-0 checkpoints selected by minimum validation loss. We use the same deterministic sample of 10,000 of the 12,000 validation examples, stratified by digit and environment. The rows show $\rvz_c$ at $\alpha=0$, $\rvz_c$ at $\alpha=192$, and $\rvz_s$ at $\alpha=192$; the columns color the same coordinates by digit and environment. At $\alpha=0$, $\rvz_c$ forms compact digit-associated groups with visible environment subdivision. At $\alpha=192$, environment colors are intermingled across much of $\rvz_c$, while digit-associated structure remains visible but incomplete. The auxiliary $\rvz_s$, directly supervised by $(y,e)$, forms compact joint groups. Each embedding is projected independently using cosine UMAP with 30 neighbors, minimum distance 0.1, and random seed 0; coordinates are therefore not aligned across rows. This nonlinear, single-seed visualization does not establish independence, factorization, or complete removal of environment information.}
    \label{fig:cmnist,umap}
\end{figure}

\begin{figure}[H]
    \centering
    \includegraphics[width=0.3\textwidth]{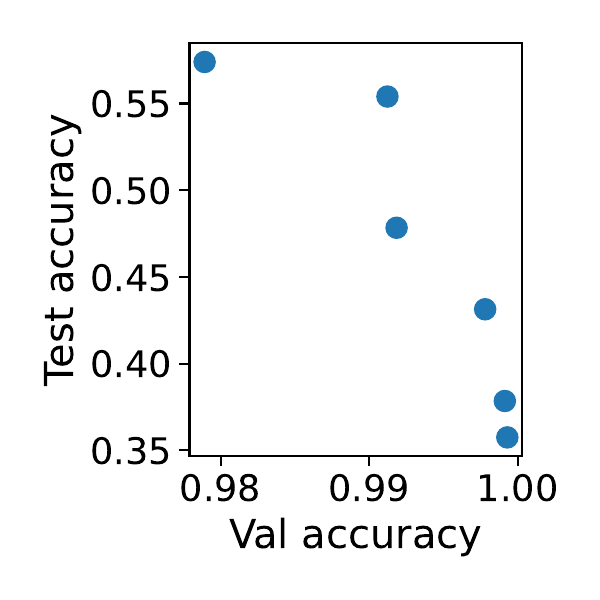}
    \caption{In- and out-of-distribution performance are negatively correlated on CMNIST, the setting in which increasing SCIL environment regularization produces the reported trade-off.}
    \label{fig:cmnist,val_test_scatter}
\end{figure}

\begin{figure}[H]
    \centering
    \includegraphics[width=0.3\textwidth]{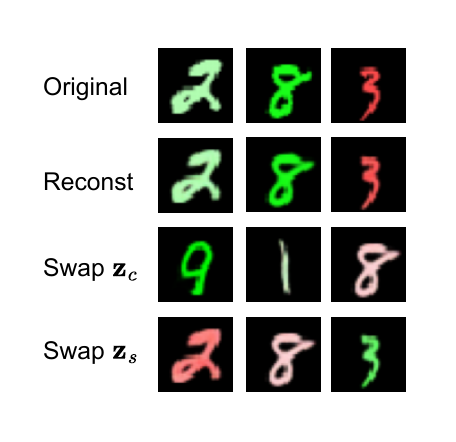}
    \caption{Qualitative iVAE decoder swaps on CMNIST. The iVAE decoder is trained jointly with its encoder, unlike the post-hoc SCIL decoder.}
    \label{fig:cmnist,gen,ivae}
\end{figure}

\begin{figure}[H]
    \centering
    \includegraphics[width=0.4\textwidth]{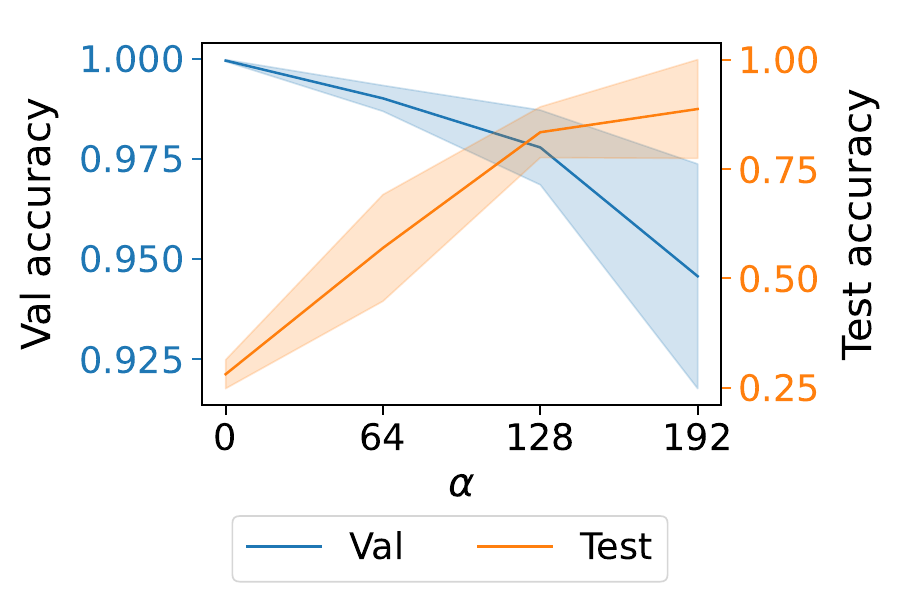}
    \caption{CMNIST $\rvz_c$-only ablation. The results are similar to the two-representation model, showing that $\rvz_s$ is not required for the downstream trade-off in this setting.}
    \label{fig:cmnist,resnet18,single_block}
\end{figure}

\begin{figure}[H]
    \centering
    \includegraphics[width=0.8\textwidth]{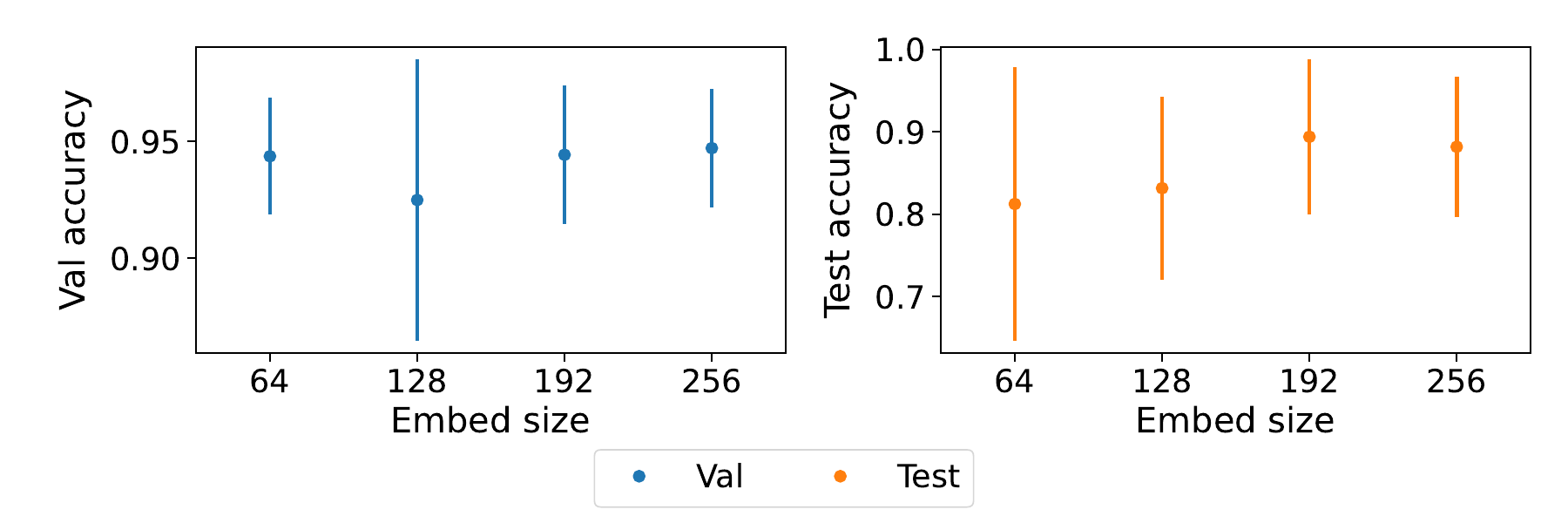}
    \caption{CMNIST embedding-size ablation for SCIL with ResNet-18 encoders and $\alpha=192$.}
    \label{fig:cmnist,z_size}
\end{figure}

\begin{figure}[H]
    \centering
    \includegraphics[width=0.8\textwidth]{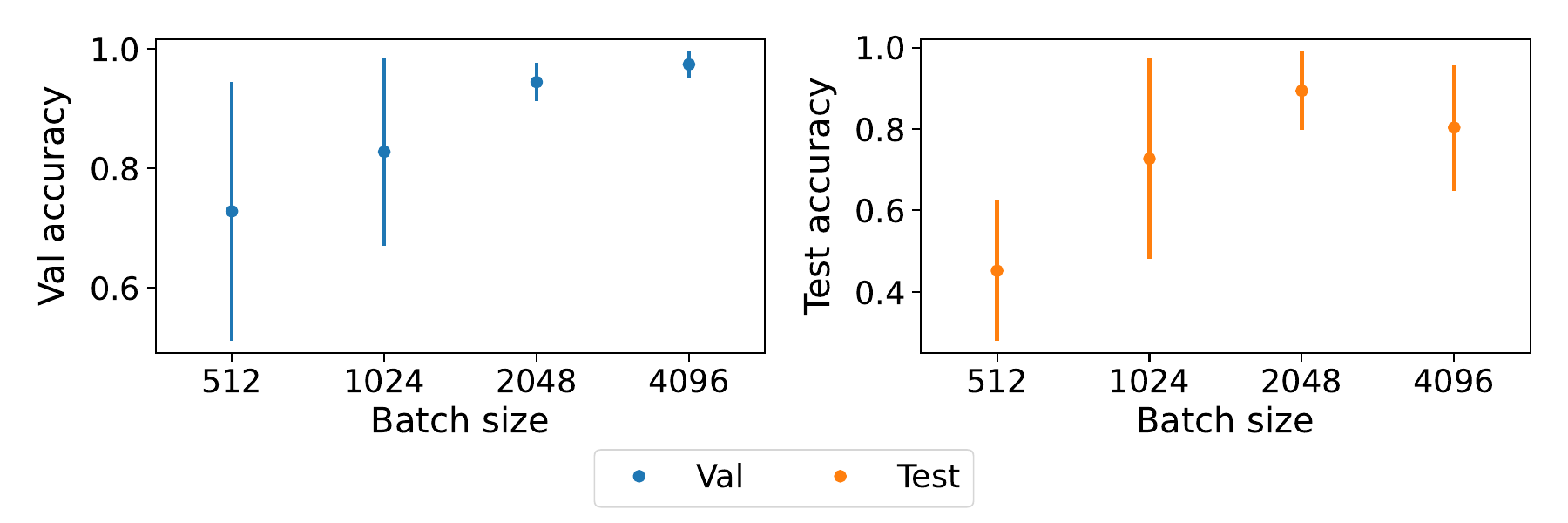}
    \caption{CMNIST batch-size ablation for SCIL with ResNet-18 encoders and $\alpha=192$. Performance is generally better with larger batches.}
    \label{fig:cmnist,batch_size}
\end{figure}

\begin{figure}[H]
    \centering
    \includegraphics[width=0.8\textwidth]{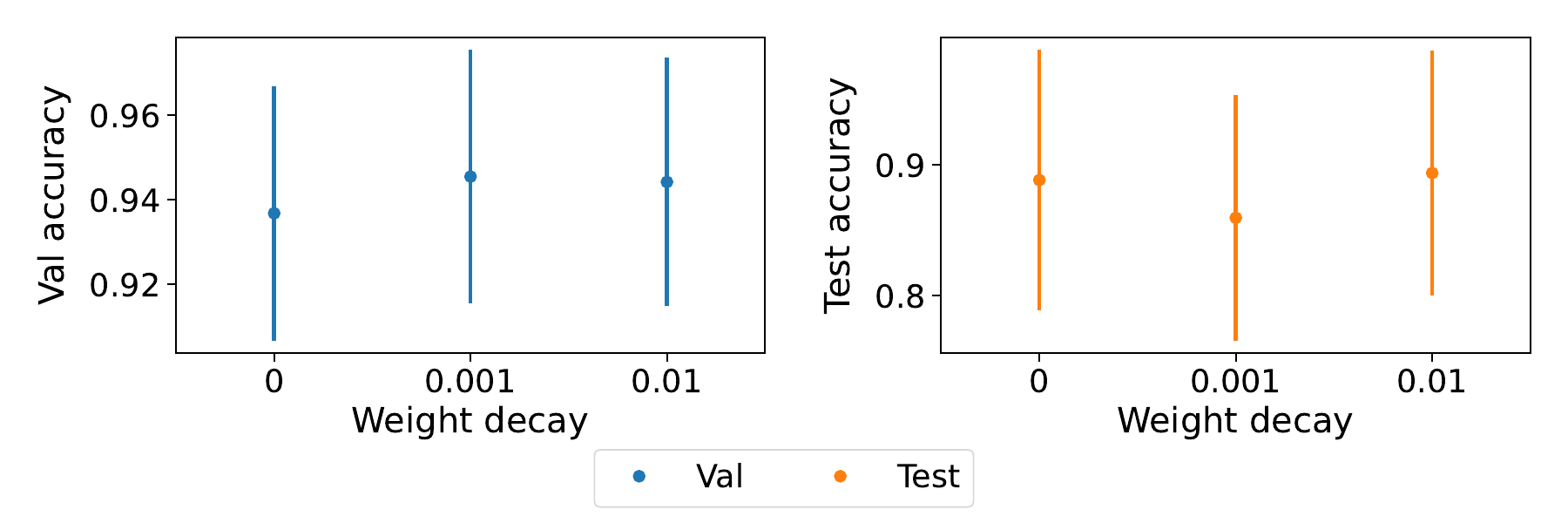}
    \caption{CMNIST weight-decay ablation for SCIL with ResNet-18 encoders and $\alpha=192$.}
    \label{fig:cmnist,weight_decay}
\end{figure}

\subsubsection{Camelyon17-WILDS}
\label{appendix:Camelyon17-WILDS}

\begin{figure}[H]
    \centering
    \includegraphics[width=0.3\textwidth]{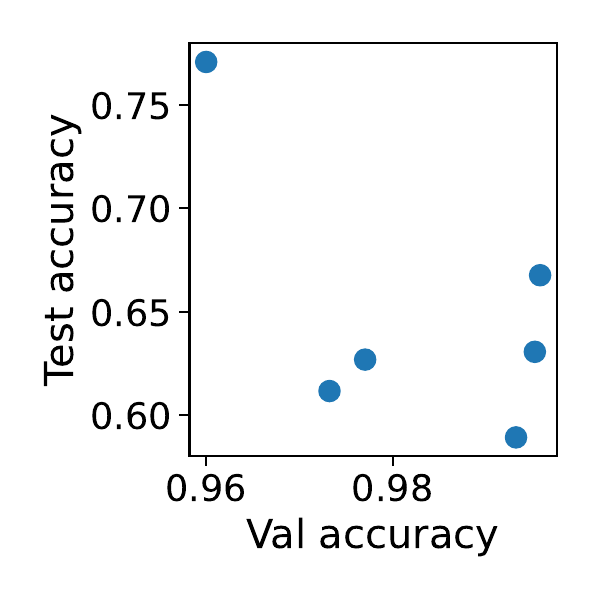}
    \caption{In- and out-of-distribution performance are negatively correlated on Camelyon17-WILDS, consistent with \citet{teney2024id}.}
    \label{fig:camelyon17,val_test_scatter}
\end{figure}

\begin{figure}[H]
    \centering
    \includegraphics[width=0.4\textwidth]{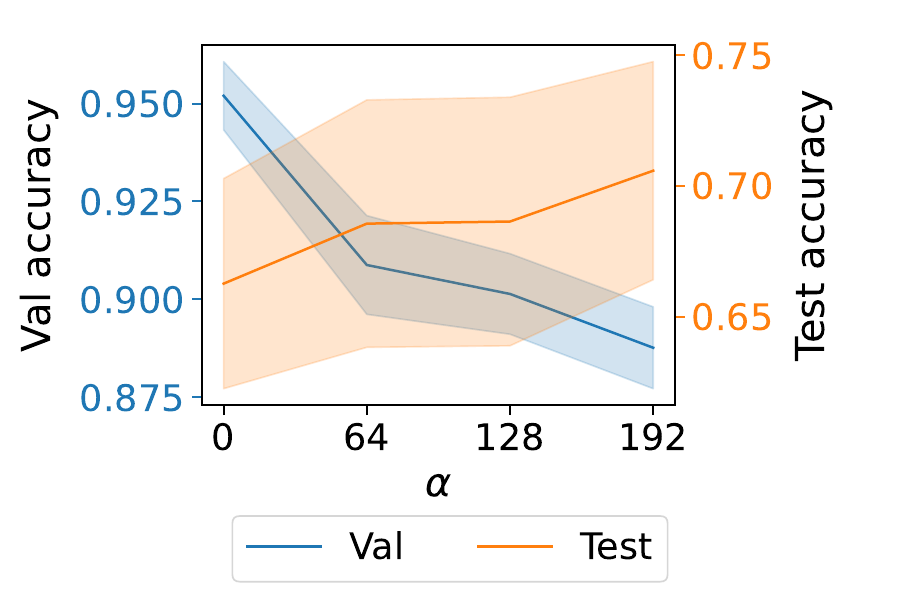}
    \caption{Camelyon17-WILDS results for SCIL using ResNet-18 encoders; the main experiment uses DenseNet-121.}
    \label{fig:camelyon17,resnet18}
\end{figure}

\begin{figure}[H]
    \centering
    \hspace*{\fill}
    \begin{subfigure}[t]{0.4\textwidth}
        \centering
        \includegraphics[width=\textwidth]{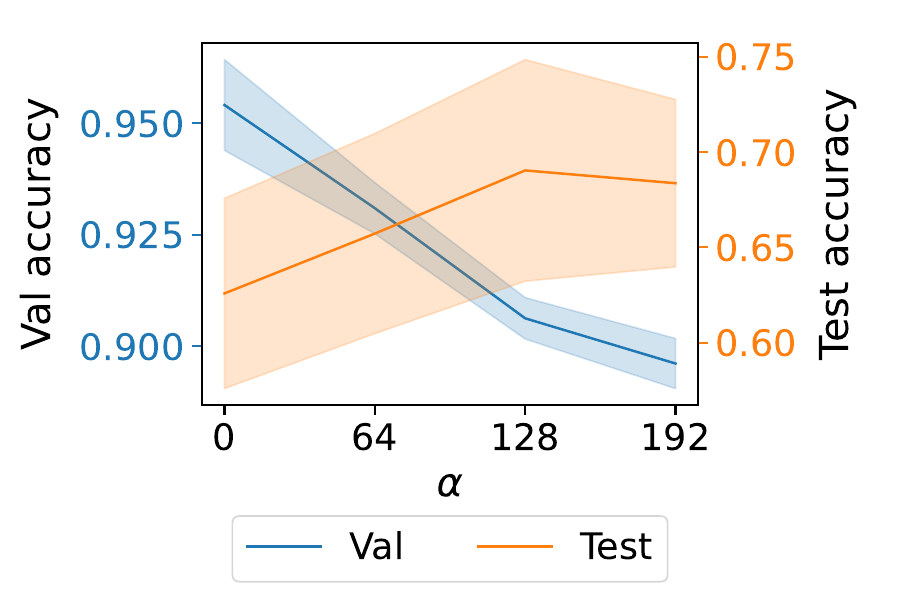}
        \caption{ResNet-18}
    \end{subfigure}
    \hfill
    \begin{subfigure}[t]{0.4\textwidth}
        \centering
        \includegraphics[width=\textwidth]{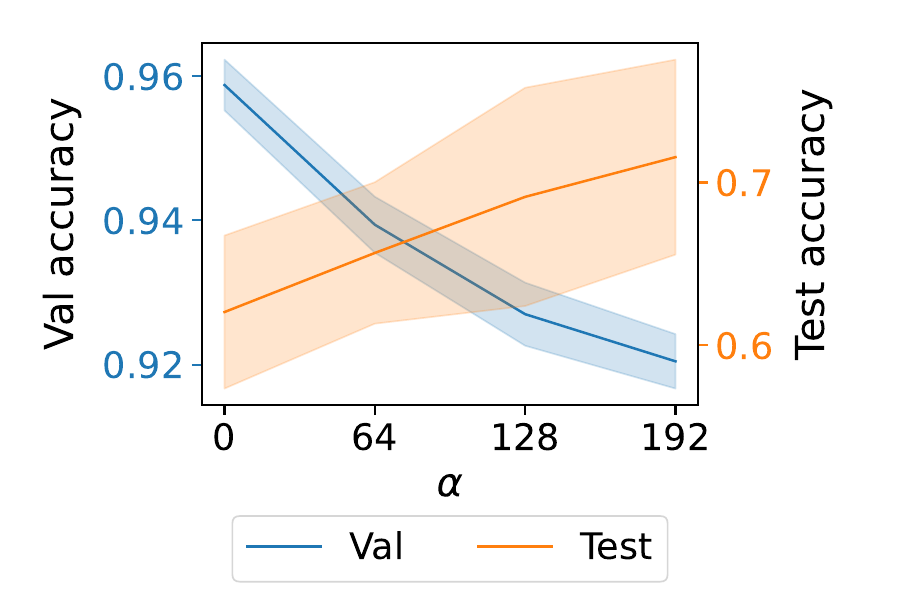}
        \caption{DenseNet-121}
    \end{subfigure}
    \hspace*{\fill}
    \caption{Camelyon17-WILDS $\rvz_c$-only ablation. Increasing $\alpha$ produces the same validation--test trade-off, with larger test-accuracy error bars than the two-representation model.}
    \label{fig:camelyon17,single_block}
\end{figure}

\begin{figure}[H]
    \centering
    \includegraphics[width=0.8\textwidth]{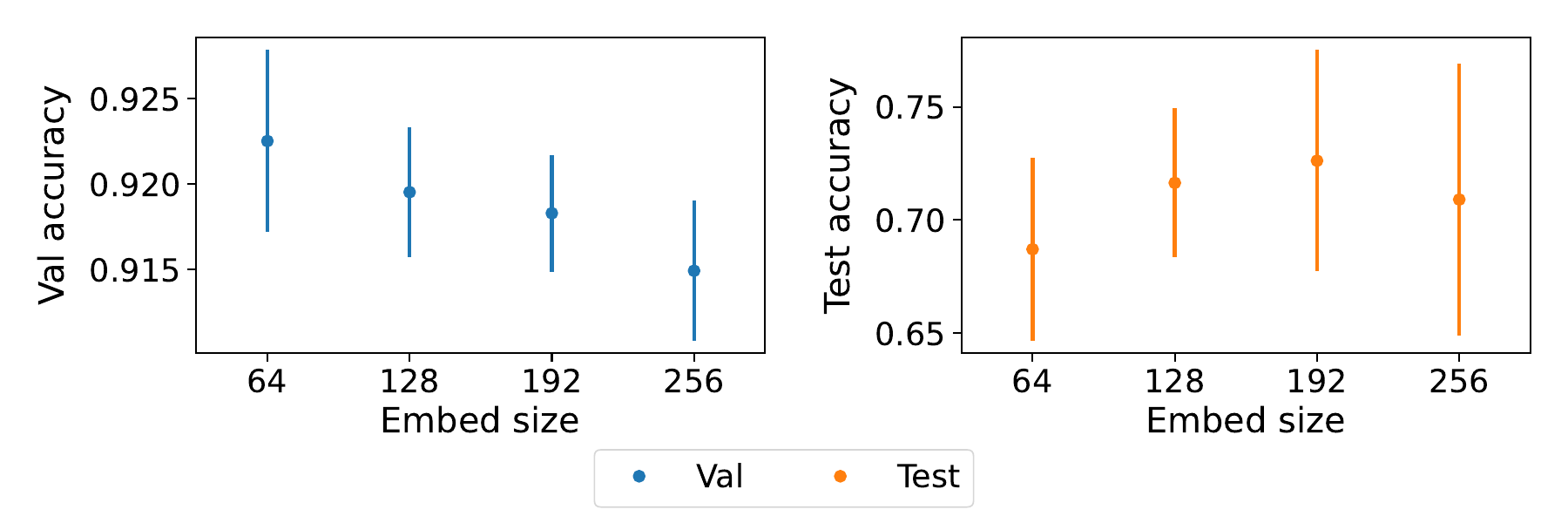}
    \caption{Camelyon17-WILDS embedding-size ablation for SCIL with DenseNet-121 encoders and $\alpha=192$.}
    \label{fig:camelyon17,z_size}
\end{figure}

\begin{figure}[H]
    \centering
    \includegraphics[width=0.8\textwidth]{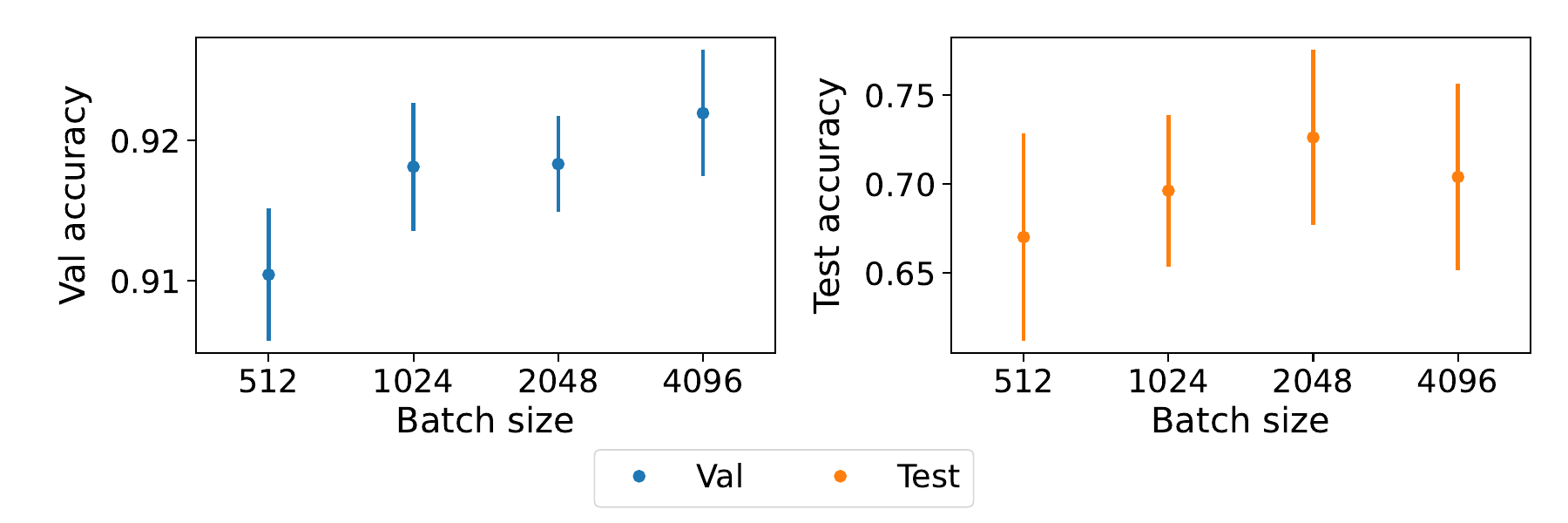}
    \caption{Camelyon17-WILDS batch-size ablation for SCIL with DenseNet-121 encoders and $\alpha=192$. Performance is generally better with larger batches.}
    \label{fig:camelyon17,batch_size}
\end{figure}

\begin{figure}[H]
    \centering
    \includegraphics[width=0.8\textwidth]{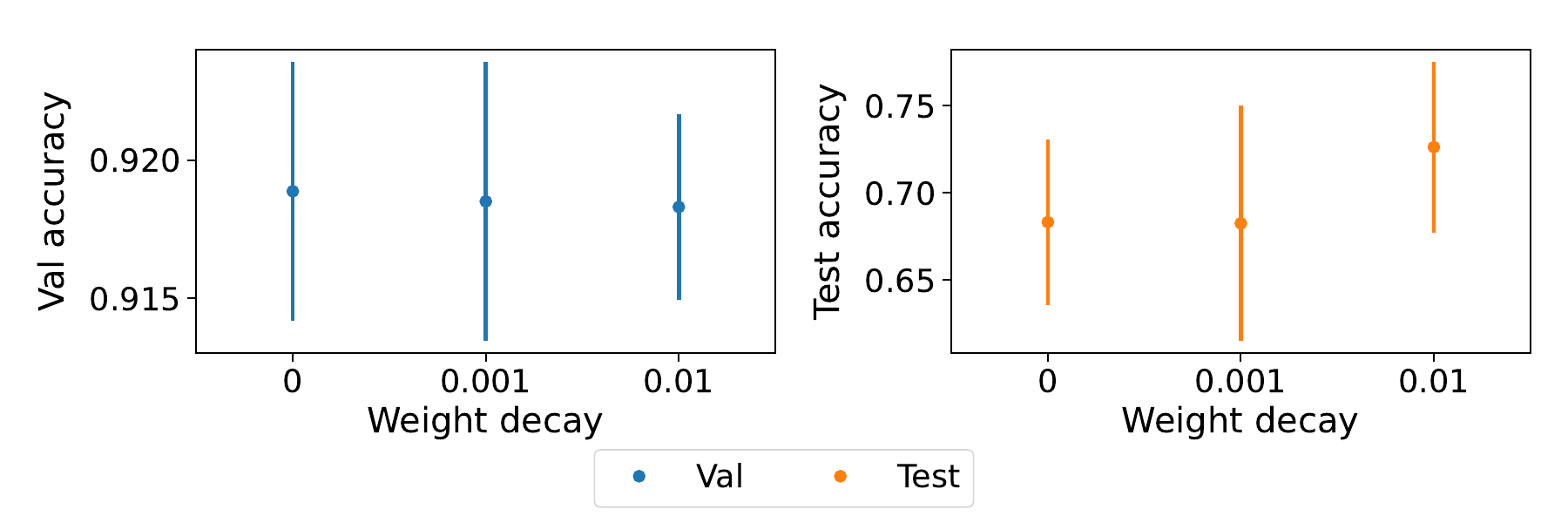}
    \caption{Camelyon17-WILDS weight-decay ablation for SCIL with DenseNet-121 encoders and $\alpha=192$.}
    \label{fig:camelyon17,weight_decay}
\end{figure}

\subsubsection{PACS}
\label{appendix:PACS}

\begin{figure}[H]
    \centering
    \hspace*{\fill}
    \begin{subfigure}[t]{0.3\textwidth}
        \centering
        \includegraphics[width=\textwidth]{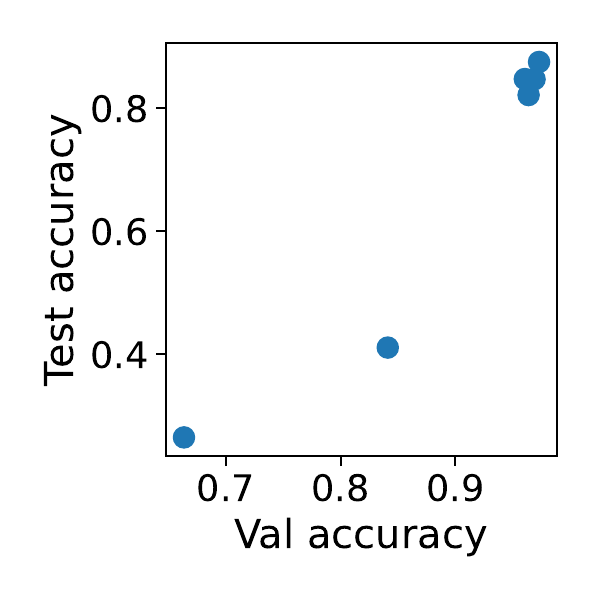}
        \caption{In- and out-of-distribution performance are positively correlated.}
    \end{subfigure}
    \hfill
    \begin{subfigure}[t]{0.4\textwidth}
        \centering
        \includegraphics[width=\textwidth]{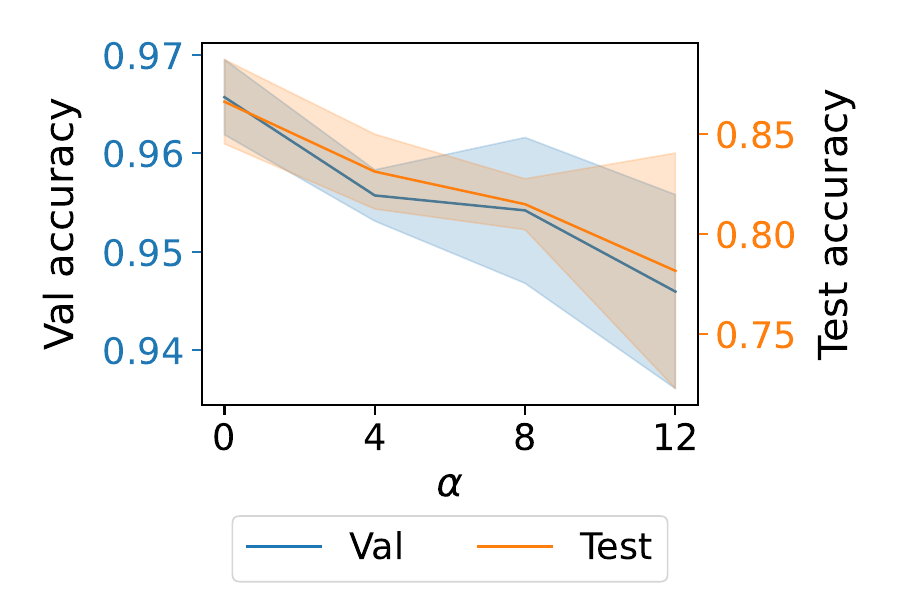}
        \caption{Increasing the environment-regularization strength $\alpha$ does not improve out-of-distribution performance.}
    \end{subfigure}
    \hspace*{\fill}
    \caption{PACS with art painting as the test domain.}
    \label{fig:PACS,art_painting}
\end{figure}

\begin{figure}[H]
    \centering
    \hspace*{\fill}
    \begin{subfigure}[t]{0.3\textwidth}
        \centering
        \includegraphics[width=\textwidth]{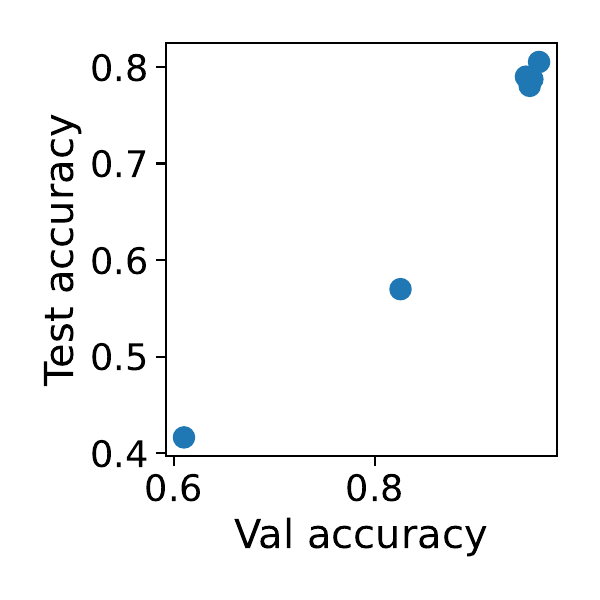}
        \caption{In- and out-of-distribution performance are positively correlated.}
    \end{subfigure}
    \hfill
    \begin{subfigure}[t]{0.4\textwidth}
        \centering
        \includegraphics[width=\textwidth]{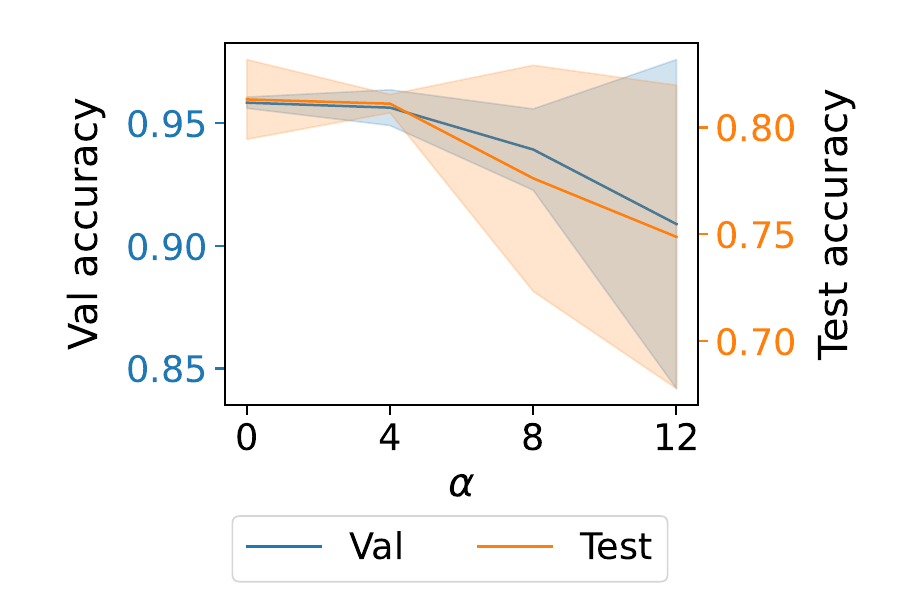}
        \caption{Increasing the environment-regularization strength $\alpha$ does not improve out-of-distribution performance.}
    \end{subfigure}
    \hspace*{\fill}
    \caption{PACS with cartoon as the test domain.}
    \label{fig:PACS,cartoon}
\end{figure}

\begin{figure}[H]
    \centering
    \hspace*{\fill}
    \begin{subfigure}[t]{0.3\textwidth}
        \centering
        \includegraphics[width=\textwidth]{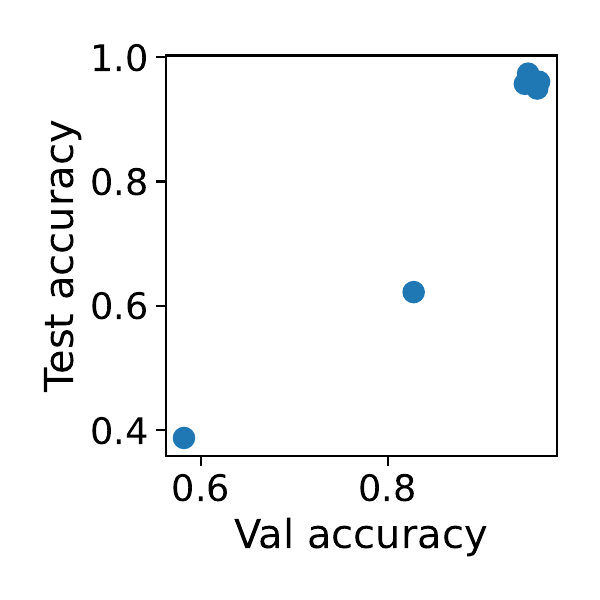}
        \caption{In- and out-of-distribution performance are positively correlated.}
    \end{subfigure}
    \hfill
    \begin{subfigure}[t]{0.4\textwidth}
        \centering
        \includegraphics[width=\textwidth]{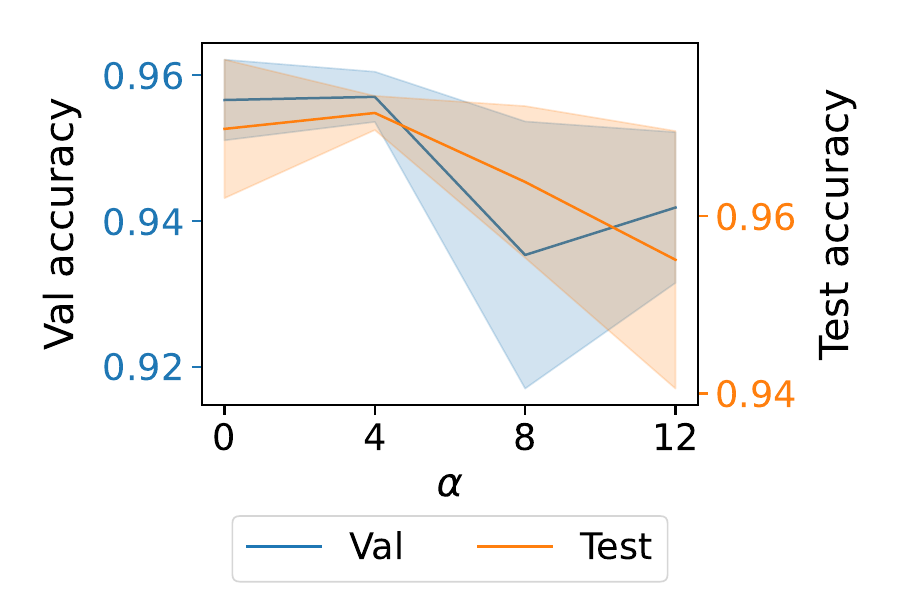}
        \caption{Increasing the environment-regularization strength $\alpha$ does not improve out-of-distribution performance.}
    \end{subfigure}
    \hspace*{\fill}
    \caption{PACS with photo as the test domain.}
    \label{fig:PACS,photo}
\end{figure}

\begin{figure}[H]
    \centering
    \hspace*{\fill}
    \begin{subfigure}[t]{0.3\textwidth}
        \centering
        \includegraphics[width=\textwidth]{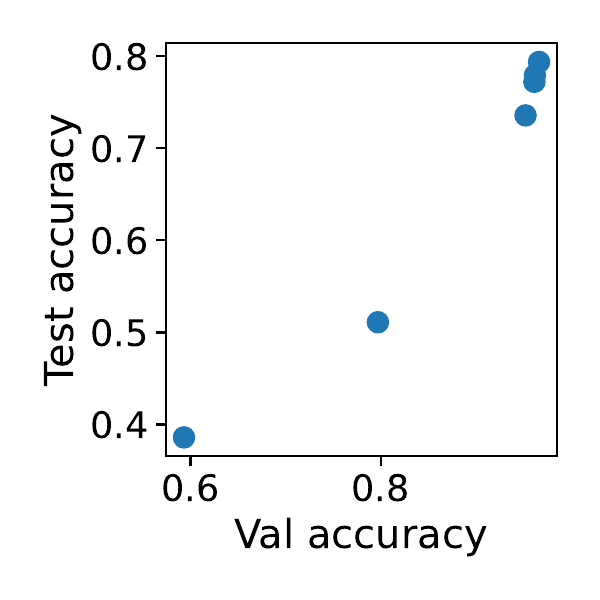}
        \caption{In- and out-of-distribution performance are positively correlated.}
    \end{subfigure}
    \hfill
    \begin{subfigure}[t]{0.4\textwidth}
        \centering
        \includegraphics[width=\textwidth]{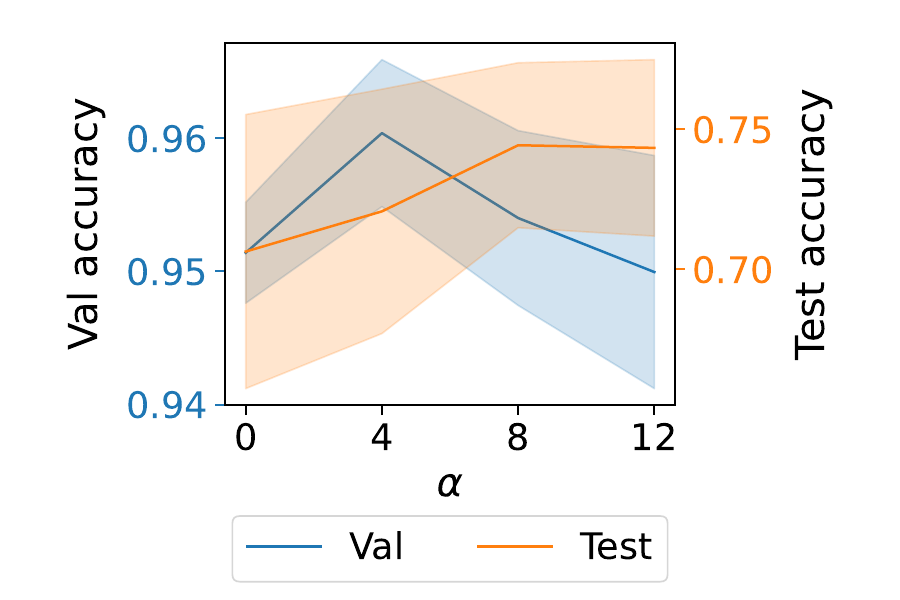}
        \caption{Increasing the environment-regularization strength $\alpha$ does not improve out-of-distribution performance.}
    \end{subfigure}
    \hspace*{\fill}
    \caption{PACS with sketch as the test domain.}
    \label{fig:PACS,sketch}
\end{figure}

\begin{table}[H]
    \centering
    \caption{Test accuracy (\%) for PACS with three random seeds.}
    \begin{tabular}{lccccc}
    \toprule
    Algorithm & Art painting & Cartoon & Photo & Sketch & Average\\
    \midrule
    SCIL ($\alpha = 0$) & $86.6 \pm 2.1$ & $81.3 \pm 1.9$ & $97.0 \pm 0.8$ & $70.6 \pm 4.9$ & $83.9$\\
    ERM & $88.1 \pm 0.1$ & $77.9 \pm 1.3$ & $97.8 \pm 0.0$ & $79.1 \pm 0.9$ & $85.7$\\
    CORAL & $87.7 \pm 0.6$ & $79.2 \pm 1.1$ & $97.6 \pm 0.0$ & $79.4 \pm 0.7$ & $86.0$\\
    DANN & $85.9 \pm 0.5$ & $79.9 \pm 1.4$ & $97.6 \pm 0.2$ & $75.2 \pm 2.8$ & $84.6$\\
    IRM & $85.0 \pm 1.6$ & $77.6 \pm 0.9$ & $96.7 \pm 0.3$ & $78.5 \pm 2.6$ & $84.4$\\
    Group DRO & $86.4 \pm 0.3$ & $79.9 \pm 0.8$ & $98.0 \pm 0.3$ & $72.1 \pm 0.7$ & $84.1$\\
    \bottomrule
    \end{tabular}
    \label{table:PACS}
\end{table}

\subsubsection{VLCS}
\label{appendix:VLCS}

\begin{figure}[H]
    \centering
    \hspace*{\fill}
    \begin{subfigure}[t]{0.3\textwidth}
        \centering
        \includegraphics[width=\textwidth]{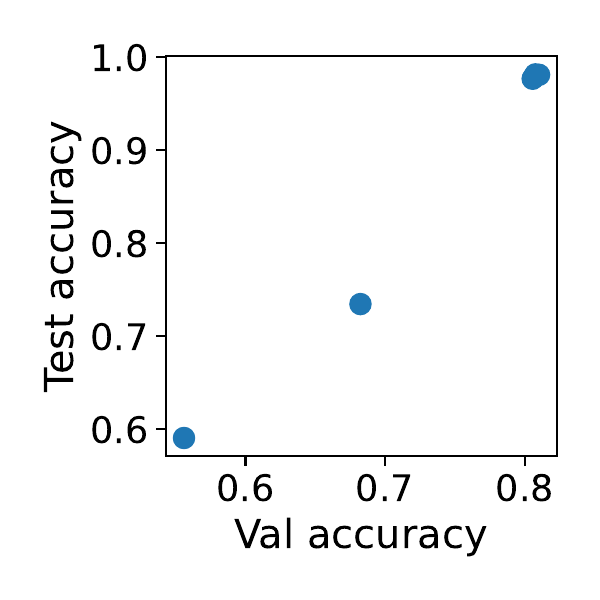}
        \caption{In- and out-of-distribution performance are positively correlated.}
    \end{subfigure}
    \hfill
    \begin{subfigure}[t]{0.4\textwidth}
        \centering
        \includegraphics[width=\textwidth]{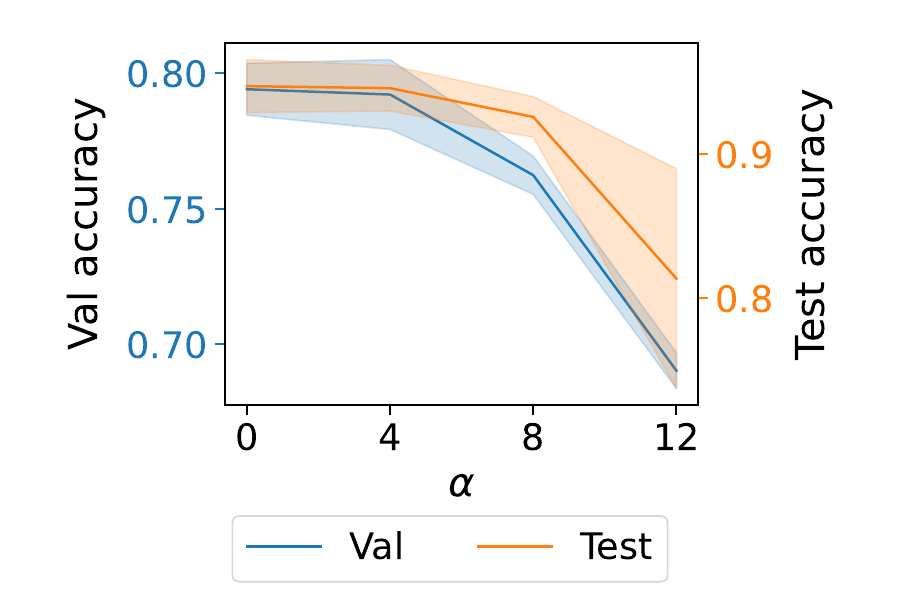}
        \caption{Increasing the environment-regularization strength $\alpha$ does not improve out-of-distribution performance.}
    \end{subfigure}
    \hspace*{\fill}
    \caption{VLCS with Caltech101 as the test domain.}
    \label{fig:VLCS,Caltech101}
\end{figure}

\begin{figure}[H]
    \centering
    \hspace*{\fill}
    \begin{subfigure}[t]{0.3\textwidth}
        \centering
        \includegraphics[width=\textwidth]{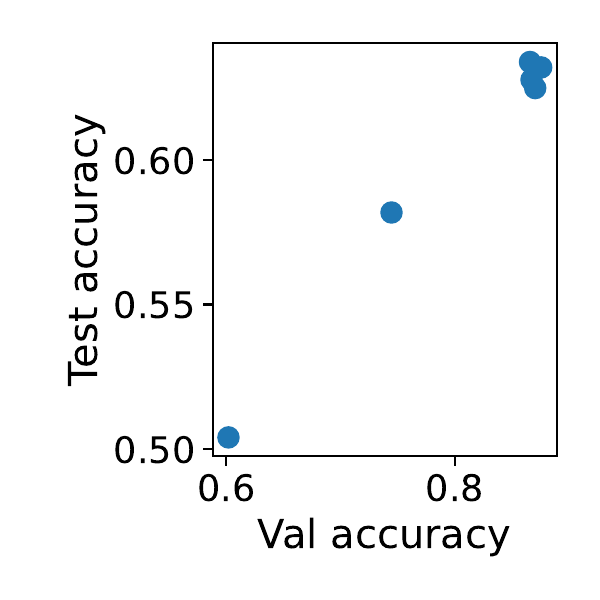}
        \caption{In- and out-of-distribution performance are positively correlated.}
    \end{subfigure}
    \hfill
    \begin{subfigure}[t]{0.4\textwidth}
        \centering
        \includegraphics[width=\textwidth]{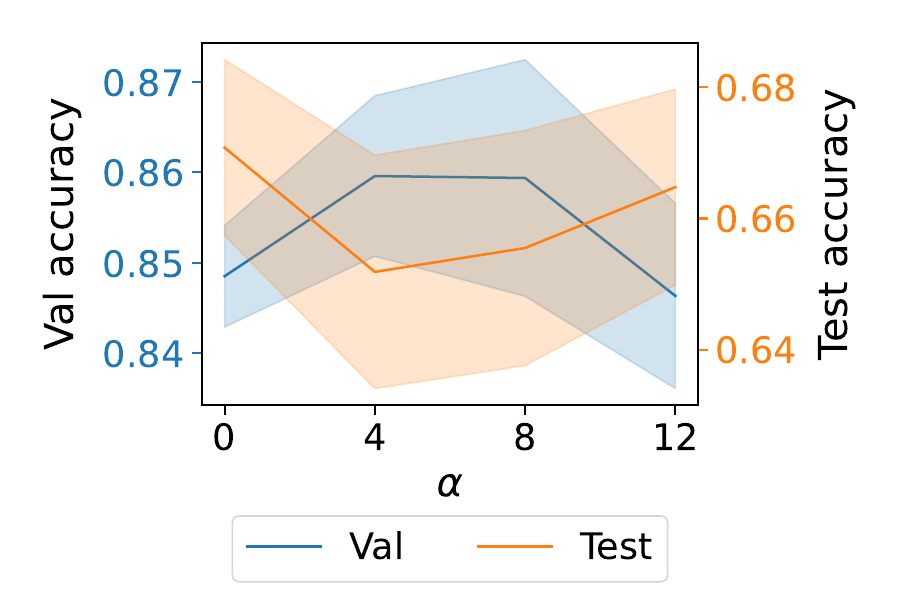}
        \caption{Increasing the environment-regularization strength $\alpha$ does not improve out-of-distribution performance.}
    \end{subfigure}
    \hspace*{\fill}
    \caption{VLCS with LabelMe as the test domain.}
    \label{fig:VLCS,LabelMe}
\end{figure}

\begin{figure}[H]
    \centering
    \hspace*{\fill}
    \begin{subfigure}[t]{0.3\textwidth}
        \centering
        \includegraphics[width=\textwidth]{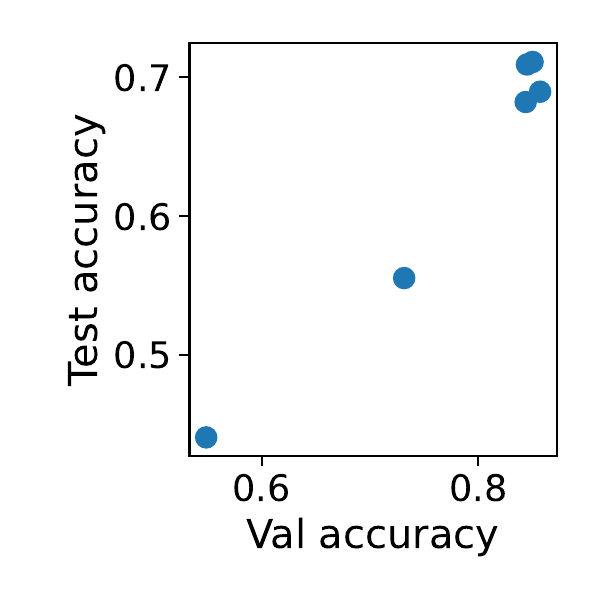}
        \caption{In- and out-of-distribution performance are positively correlated.}
    \end{subfigure}
    \hfill
    \begin{subfigure}[t]{0.4\textwidth}
        \centering
        \includegraphics[width=\textwidth]{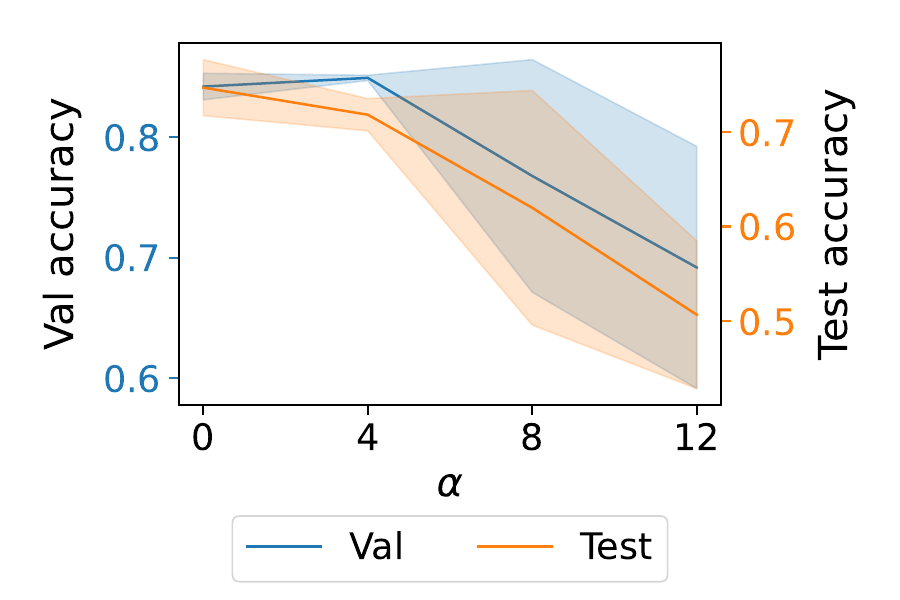}
        \caption{Increasing the environment-regularization strength $\alpha$ does not improve out-of-distribution performance.}
    \end{subfigure}
    \hspace*{\fill}
    \caption{VLCS with SUN09 as the test domain.}
    \label{fig:VLCS,SUN09}
\end{figure}

\begin{figure}[H]
    \centering
    \hspace*{\fill}
    \begin{subfigure}[t]{0.3\textwidth}
        \centering
        \includegraphics[width=\textwidth]{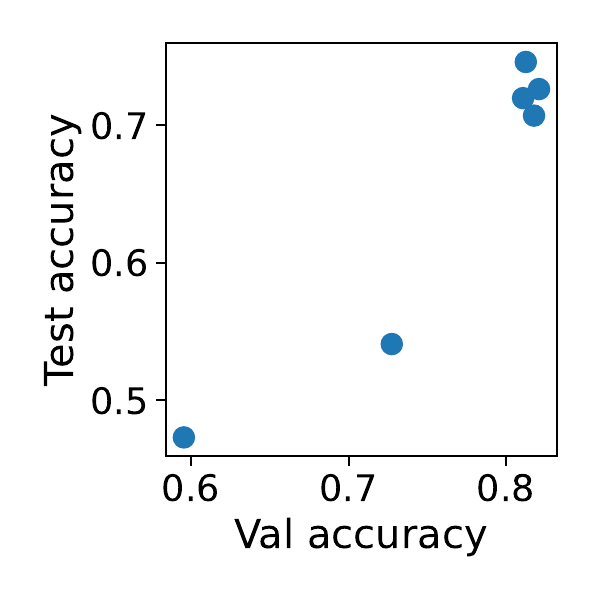}
        \caption{In- and out-of-distribution performance are positively correlated.}
    \end{subfigure}
    \hfill
    \begin{subfigure}[t]{0.4\textwidth}
        \centering
        \includegraphics[width=\textwidth]{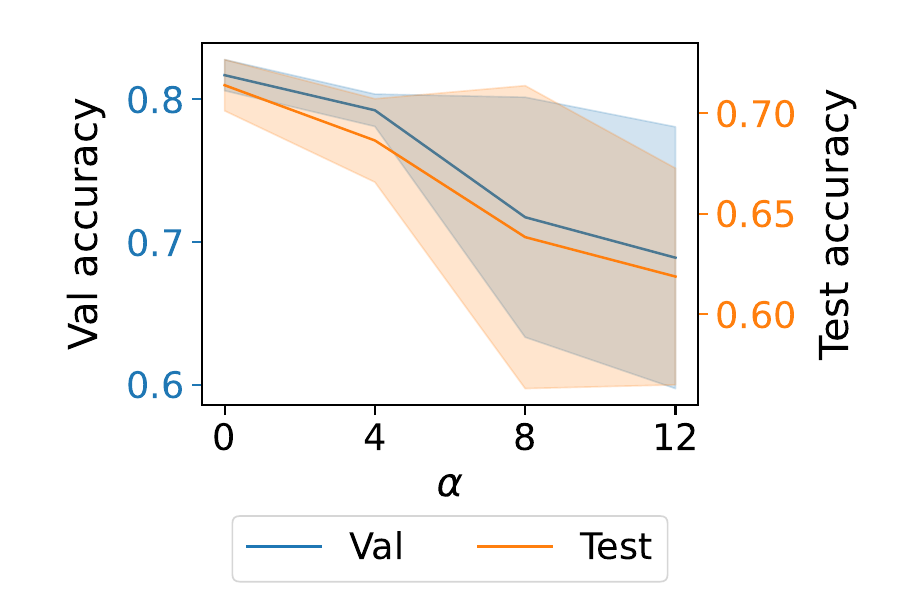}
        \caption{Increasing the environment-regularization strength $\alpha$ does not improve out-of-distribution performance.}
    \end{subfigure}
    \hspace*{\fill}
    \caption{VLCS with VOC2007 as the test domain.}
    \label{fig:VLCS,VOC2007}
\end{figure}

\begin{table}[H]
    \centering
    \caption{Test accuracy (\%) for VLCS with three random seeds.}
    \begin{tabular}{lccccc}
    \toprule
    Algorithm & Caltech101 & LabelMe & SUN09 & VOC2007 & Average\\
    \midrule
    SCIL ($\alpha = 0$) & $94.7 \pm 1.8$ & $67.1 \pm 1.3$ & $74.7 \pm 3.0$ & $71.4 \pm 1.3$ & $77.0$\\
    ERM & $97.6 \pm 1.0$ & $63.3 \pm 0.9$ & $72.2 \pm 0.5$ & $76.4 \pm 1.5$ & $77.4$\\
    CORAL & $98.8 \pm 0.1$ & $64.6 \pm 0.8$ & $71.7 \pm 1.4$ & $75.8 \pm 0.4$ & $77.7$\\
    DANN & $98.5 \pm 0.2$ & $64.9 \pm 1.1$ & $73.1 \pm 0.7$ & $78.3 \pm 0.3$ & $78.7$\\
    IRM & $97.6 \pm 0.3$ & $65.0 \pm 0.9$ & $72.2 \pm 0.5$ & $76.4 \pm 1.5$ & $78.1$\\
    Group DRO & $97.7 \pm 0.4$ & $62.5 \pm 1.1$ & $70.1 \pm 0.7$ & $78.4 \pm 0.9$ & $77.2$\\
    \bottomrule
    \end{tabular}
    \label{table:VLCS}
\end{table}

\subsection{Batch correction}
\label{appendix:batch correction}

\paragraph{CORUM prediction}

Our CORUM prediction task follows \citet{wang2023multi}, with modifications for comparison with CellProfiler. We compute $\rvz_c$ for every cell in the training, validation, and test sets and retain cells with corresponding CellProfiler features. We average cell embeddings within each sgRNA, then average the sgRNA embeddings for each gene. Finally, we subtract the non-targeting-control embedding and standardize each of the 64 dimensions to zero mean and unit variance.

Then, we incorporate the CORUM database, which defines the pairs of genes that belong to the same protein complex. We discard all gene embeddings that are not in this database. We compute the cosine similarity between each pair of gene embeddings, and interpret it as the prediction that they belong to the same family. The prediction target is one if they belong to the same family according to the CORUM database, and zero otherwise. We turn the cosine similarities into binary predictions by across various prediction thresholds by using the $i$'th percentile as the upper threshold and the $100 - i$'th percentile as the lower threshold for each integer $i \in \{80, \dotsc, 100\}$. Finally, we use the binary predictions and prediction targets to obtain a precision and recall at each value of $i$, and plot the precision and recall curve.

\paragraph{Batch prediction}

For well prediction, we compute $\rvz_c$ for examples from the training, validation, and test sets and retain examples with corresponding CellProfiler features. We randomly use 60\% of these embeddings to fit logistic regression, evaluate on the remaining 40\%, and report F1.

\section{Code}
\label{appendix:code}

A PyTorch implementation of the supervised contrastive and environment-regularization losses is provided below.
\begin{verbatim}
def supcon_loss(z, u, temperature):
    batch_size = len(z)
    u_col = u.unsqueeze(1)
    u_row = u.unsqueeze(0)
    mask_pos = (u_col == u_row).float()
    offdiag_mask = 1. - torch.eye(batch_size)
    mask_pos = mask_pos * offdiag_mask
    logits = torch.matmul(z, z.T) / temperature
    p = mask_pos / mask_pos.sum(dim=1, keepdim=True).clamp(min=1.)
    q = F.log_softmax(logits, dim=1)
    return F.cross_entropy(q, p)

def invariance_loss(zc, e, temperature):
    batch_size = len(zc)
    e_col = e.unsqueeze(1)
    e_row = e.unsqueeze(0)
    mask_pos = (e_col == e_row).float()
    mask_neg = 1. - mask_pos
    offdiag_mask = 1. - torch.eye(batch_size)
    mask_pos = mask_pos * offdiag_mask
    logits = torch.matmul(zc, zc.T) / temperature
    q = F.log_softmax(logits, dim=1)
    log_prob_pos = (q * mask_pos).mean(dim=1)
    log_prob_neg = (q * mask_neg).mean(dim=1)
    return (log_prob_pos - log_prob_neg).abs().mean()
\end{verbatim}

\end{document}